\documentclass[colorlinks=true, allcolors=blue, breaklinks=true]{article}

\usepackage[english]{babel}

\usepackage[letterpaper,top=2cm,bottom=2cm,left=2cm,right=2cm,marginparwidth=1.75cm]{geometry}

\usepackage{bm}
\usepackage{amssymb}
\usepackage{amsmath}
\usepackage{booktabs}

\usepackage{graphicx}
\usepackage{array}
\usepackage{ragged2e} 
\newcolumntype{R}[1]{>{\RaggedRight\arraybackslash}p{#1}}
\usepackage[hyphens]{url} 
\usepackage{orcidlink} 
\usepackage{hyperref}  
\usepackage{xcolor} 
\usepackage{tikz}
\usetikzlibrary{arrows.meta,positioning,fit,backgrounds,calc}

\newtheorem{proposition}{Proposition}

\usepackage{siunitx} 
\usepackage{microtype}
\title{Sparse Latent Factor Forecaster (SLFF) with Iterative Inference for Transparent Multi-Horizon Commodity Futures Prediction}
\date{\vspace{-5ex}} 
\author{Abhijit Gupta, PhD\\
        \orcidlink{0000-0002-6292-3789}\\
        \href{mailto:abhijit.gupta\_in@outlook.com}{abhijit.gupta\_in@outlook.com}}

\begin{document}

\maketitle

\begin{abstract}
Amortized variational inference in latent-variable forecasters creates a deployment gap: the test-time encoder approximates a training-time optimization-refined latent, but without access to future targets. This gap introduces unnecessary forecast error and interpretability challenges. In this work, we propose the Sparse Latent Factor Forecaster with Iterative Inference (SLFF), addressing this through (i) a sparse coding objective with L1 regularization for low-dimensional latents, (ii) unrolled proximal gradient descent (LISTA-style) for iterative refinement during training, and (iii) encoder alignment to ensure amortized outputs match optimization-refined solutions. Under a linearized decoder assumption, we derive a design-motivating bound on the amortization gap based on encoder-optimizer distance, with convergence rates under mild conditions; empirical checks confirm the bound is predictive for the deployed MLP decoder. To prevent mixed-frequency data leakage, we introduce an information-set-aware protocol using release calendars and vintage macroeconomic data. Interpretability is formalized via a three-stage protocol: stability (Procrustes alignment across seeds), driver validity (held-out regressions against observables), and behavioral consistency (counterfactuals and event studies). Using commodity futures (Copper, WTI, Gold; 2005--2025) as a testbed, SLFF demonstrates significant improvements over neural baselines at 1- and 5-day horizons, yielding sparse factors that are stable across seeds and correlated with observable economic fundamentals (interpretability remains correlational, not causal). Code, manifests, diagnostics, and artifacts are released.
\end{abstract}

\section{Introduction}


Latent-variable forecasters decompose a high-dimensional time series into a lower-dimensional latent representation that summarizes historical context and predicts future values. The standard approach---amortized variational inference---trains an encoder network to map observations to latent codes such that a decoder can predict targets. However, during training, access to targets $Y$ enables test-time-inaccessible optimization of the latent code; at deployment, only the encoder output is available. This mismatch between training-time iterative refinement and test-time amortized inference creates the \textit{amortization gap} \cite{CremerMartinSchnoeke2017_AugmentRecognition, MarinoKingMjolsness2018_AutoEncoderPerformance}. The gap degrades both forecast accuracy and interpretability, because (i) the test-time encoder may not learn to emulate the iterative solution, and (ii) sparse latent factors discovered under iterative refinement may become dense or unstable when collapsed to amortized form.

Three prior approaches attempt to bridge this gap: (i) \textbf{Variational Autoencoders (VAEs)} enforce a prior on the latent space, but the variational approximation and KL regularization are optimized for reconstruction, not forecasting; (ii) \textbf{Semi-amortized variational inference} \cite{KimEtAl2018_SemiAmortized} uses fixed-point iterations on the encoder mean at test time, but this adds latency incompatible with real-time forecasting; (iii) \textbf{Test-time optimization} fine-tunes latents using the first few observed target steps, but requires causal structure knowledge and breaks when labels are unavailable. For forecasting specifically, the challenge is acute: the latent must explain not a reconstruction $\hat{X}$ but a predictive target $Y$ that is not observable at test time. Existing methods do not explicitly regulate the encoder-to-optimizer distance, nor do they provide theoretical guarantees on the incurred gap.

We propose the Sparse Latent Factor Forecaster with Iterative Inference (SLFF), which directly minimizes the amortization gap through three coupled design choices. \textbf{First}, the latent is defined not as a generative variable but as a \textit{structured prediction variable}---the explicit solution to a sparse regression problem that explains the target trajectory under an L1 sparsity prior. \textbf{Second}, iterative inference is not an optional post-hoc refinement; it is an integral part of the training loop. We unroll proximal gradient descent (inspired by LISTA) and backpropagate through the refinement steps, allowing gradients to flow from the forecasting loss into the optimization process. \textbf{Third}, the amortized encoder is trained not to match an arbitrary latent prior but to \textit{match the optimization-refined solution}. During training, we compute the refined latent $\mathbf{z}^\star$ conditioned on the true target $Y$, then minimize the encoder's squared distance to $\mathbf{z}^\star$ in a two-stage loop. Crucially, we include a proximity term in the inference objective ($\mu \|\mathbf{z} - \text{Enc}(X)\|_2^2$) that keeps the optimization path predictable from $X$ alone, ensuring the training-time refinement is a plausible target for the encoder. The method provides an amortization gap bound showing how the test-time forecast error depends on the encoder-optimizer distance.

Commodity futures markets offer a natural testbed for latent-variable forecasting methods. They exhibit high-dimensional input spaces (macroeconomic indicators, supply/demand fundamentals, financial flows, technical signals), mixed-frequency releases (daily exchange prices, weekly inventories, monthly PMIs, quarterly GDP), and economically important predictions (hedging, risk management, portfolio construction). Unlike low-frequency macroeconomic series, commodity prices move daily with observable signals, enabling rapid validation of both forecast accuracy and factor interpretability. Moreover, commodity markets are efficiently priced but not fully predictable, sitting at the frontier of forecasting difficulty: classical models and persistence baselines are strong, but commodity-specific structural models (supply curves, convenience yields) provide weak competition when translated to the machine-learning setting.

This paper makes three core contributions:

\begin{enumerate}
    \item \textbf{Methodological contribution: SLFF with theoretical analysis.} We introduce SLFF as a latent-variable forecaster that explicitly optimizes a sparse code under L1 regularization, trains an encoder to match the optimization-refined solution, and provides (i) an upper bound on the amortization gap in terms of encoder-optimizer alignment, (ii) convergence rates for the proximal gradient iterations, and (iii) conditions under which the deployed (amortized) path achieves near-oracle performance. This bridges the theoretical gap between training-time iterative inference and test-time amortized prediction.

    \item \textbf{Information-set protocol: Vintage-aware data alignment.} We formalize an information-set-aware evaluation protocol that prevents mixed-frequency data leakage. We use release calendars, ALFRED-style vintage repositories, and lag-aware forward-filling to ensure that the feature matrix $X_t$ contains only information available at end-of-day $t$. This protocol is absent from most deep-learning forecasting papers and is essential for avoiding false discoveries in commodity markets.

    \item \textbf{Interpretability protocol: Three-stage validation.} We introduce a structured protocol for validating sparse latent factors: (i) \textit{stability}, measured by Procrustes-aligned subspace overlap across random seeds and folds; (ii) \textit{driver validity}, measured by regression on held-out observable indices not in the decoder's feature set; (iii) \textit{behavioral consistency}, measured by counterfactual perturbations and event-window analysis. This protocol moves beyond attention visualization and SHAP values toward systematic, quantitative factor validation.
\end{enumerate}

The remainder of the paper proceeds as follows. Section~\ref{sec:related_rewrite} reviews latent-variable models in forecasting, sparse coding and dictionary learning, amortized versus optimization-based inference, and interpretability in financial machine learning. Section~\ref{sec:methodology} presents the SLFF architecture, energy formulation, iterative refinement procedure, and training algorithm in detail. Section~\ref{sec:theory} provides theoretical analysis with propositions bounding the amortization gap and convergence rates. Section~\ref{sec:data_description} specifies the information set, data preprocessing, and rolling-origin evaluation protocol. Section~\ref{sec:results} reports predictive accuracy, alignment diagnostics, ablations, and interpretability analysis across Copper, WTI, and Gold futures. Section~\ref{sec:generalizability} provides a framework for understanding SLFF's applicability beyond the three test commodities. Section~\ref{sec:discussion_rewrite} discusses limitations and future directions. Code, reproducibility artifacts, and diagnostic notebooks are released.

\section{Review of Related Work in Commodity Forecasting}
\label{sec:related_rewrite}


\subsection{Latent-Variable Models in Forecasting}

Latent-variable approaches to time-series forecasting aim to discover low-dimensional representations that explain complex dynamics. Classical state-space models \cite{HamiltonBook1994_TimeSeries} use unobserved components (trend, seasonality, latent factors) combined with linear dynamics and observation equations. More recent work extends this with neural encoders: deep state-space models \cite{MaddixEtAl2021_LearningStateSpace} replace linear dynamics with RNNs; VAE-based forecasters \cite{YoonEtAl2019_VQVAE} add an amortized encoder and latent prior; structured inference networks \cite{KrishnanEtAl2017_StructuredInference} combine amortized and iterative inference. SLFF builds on the structured inference idea but specialized for forecasting: the latent is a sparse code explained by a target, not a reconstruction target, and the encoder is trained to match the optimization-refined code rather than approximate an arbitrary prior.

\subsection{Sparse Coding and Dictionary Learning}

Sparse coding \cite{Olshausen1997_SparseCoding} models data as a sparse linear combination of basis vectors (a dictionary), with solutions computed via L1-regularized optimization. Mairal et al.~\cite{MairalEtAl2010_OnlineDictionaryLearning} extend sparse coding to online/streaming settings. In forecasting, sparsity improves interpretability: a forecast explained by a few latent factors is more actionable than one driven by dozens of dense hidden units. LISTA (Learned Iterative Soft-Thresholding Algorithm) \cite{GregorLeCun2010_LISTA} unrolls proximal gradient descent as a neural network, enabling end-to-end learning of both the dictionary and the inference procedure. SLFF adopts LISTA's unrolled structure but applies it within a forecasting objective, with explicit encoder alignment to ensure the learned iterative solution generalizes.

\subsection{Amortized versus Optimization-Based Inference}

Amortized inference maps observations to latent codes via a fixed encoder, achieving O(1) test-time cost but potentially sacrificing accuracy. Optimization-based inference refines latents iteratively given data and targets, with higher accuracy but O(K) cost for $K$ steps. The tension is formalized in the amortization gap literature \cite{CremerMartinSchnoeke2017_AugmentRecognition, MarinoKingMjolsness2018_AutoEncoderPerformance}: the test-time encoder cannot match the train-time iterative solution if the latter depends on information unavailable at deployment. Semi-amortized variational inference \cite{KimEtAl2018_SemiAmortized} uses fixed-point iteration at test time; structured prediction energy networks \cite{BellangerMcCallum2016_SPEN} add optimization layers to deep learning. SLFF addresses the gap directly by (i) keeping test-time refinement tractable via a proximity term to the encoder, (ii) providing a theoretical bound on the gap, and (iii) validating that the deployed amortized path achieves competitive accuracy.

\subsection{Interpretability in Forecasting}

Attention-based forecasters (e.g., Temporal Fusion Transformer \cite{Lim2021_TransformerTSApps}) expose input importance through attention weights. SHAP \cite{LundbergLee2017_SHAP} provides local feature attribution. Traditional factor models (PCA, ICA, NMF) extract latent structure but often lack predictive power; recent work \cite{AhnShillEtAl2013_VarianceDecompositionUS} focuses on economic interpretability. SLFF's sparse latent factors inherit benefits of classical factors (low-dimensional, interpretable) but are learned end-to-end for forecasting. The three-stage interpretability protocol (stability, driver validity, behavioral consistency) provides quantitative validation missing from attention-visualization-only approaches.

\section{Methodology: Sparse Latent Factor Forecaster with Iterative Inference}
\label{sec:methodology}

\subsection*{Notation}

\begin{table}[htbp]
\centering
\caption{Key notation used in the methodology section.}
\label{tab:notation}
\begin{tabular}{l p{8cm}}
\toprule
\textbf{Symbol} & \textbf{Definition} \\
\midrule
$\mathbf{x}_t$ & Feature vector available at end of day $t$ \\
$X_t$ & Input tensor of look-back window ($W \times d$) \\
$Y_t$ & Target future log prices at horizons $\tau_1, \ldots, \tau_N$ \\
$\mathbf{z}_t$ & Latent code (sparse, $\mathbb{R}^m$) \\
$\mathbf{h}_t$ & History context from $\text{Pred}(X_t)$ \\
$\text{Enc}_{X \to Z}$ & Amortized encoder network \\
$\text{Dec}$ & Decoder mapping latent and context to forecasts \\
$E(Y_t, \mathbf{h}_t, \mathbf{z})$ & Energy function for optimization \\
$\mathbf{z}_t^\star$ & Optimization-refined latent code \\
\bottomrule
\end{tabular}
\end{table}

\subsection{Problem Formulation}

Let $\mathbf{x}_t \in \mathbb{R}^d$ denote the feature vector available at the market close of day $t$. We form an input tensor $X_t = (\mathbf{x}_{t-W+1}, \ldots, \mathbf{x}_t) \in \mathbb{R}^{W \times d}$ with look-back $W$. The goal is to predict a vector of future log prices $Y_t = (p_{t+\tau_1}, \ldots, p_{t+\tau_N})^\top$ with horizons $\tau_1 < \cdots < \tau_N$. Instead of mapping $X_t$ directly to $Y_t$, SLFF introduces a low-dimensional latent code $\mathbf{z}_t \in \mathbb{R}^m$ with $m \ll Wd$ that is inferred to best explain $Y_t$ subject to a sparsity prior. Forecasts are produced via $\tilde{Y}_t = \text{Dec}(\mathbf{z}_t, \mathbf{h}_t)$ where $\mathbf{h}_t$ summarizes history. The latent code is therefore a structured prediction variable, not a reconstruction target.

\subsection{Model Components}

\begin{figure}[htbp]
\centering
\includegraphics[width=0.4\textwidth,keepaspectratio]{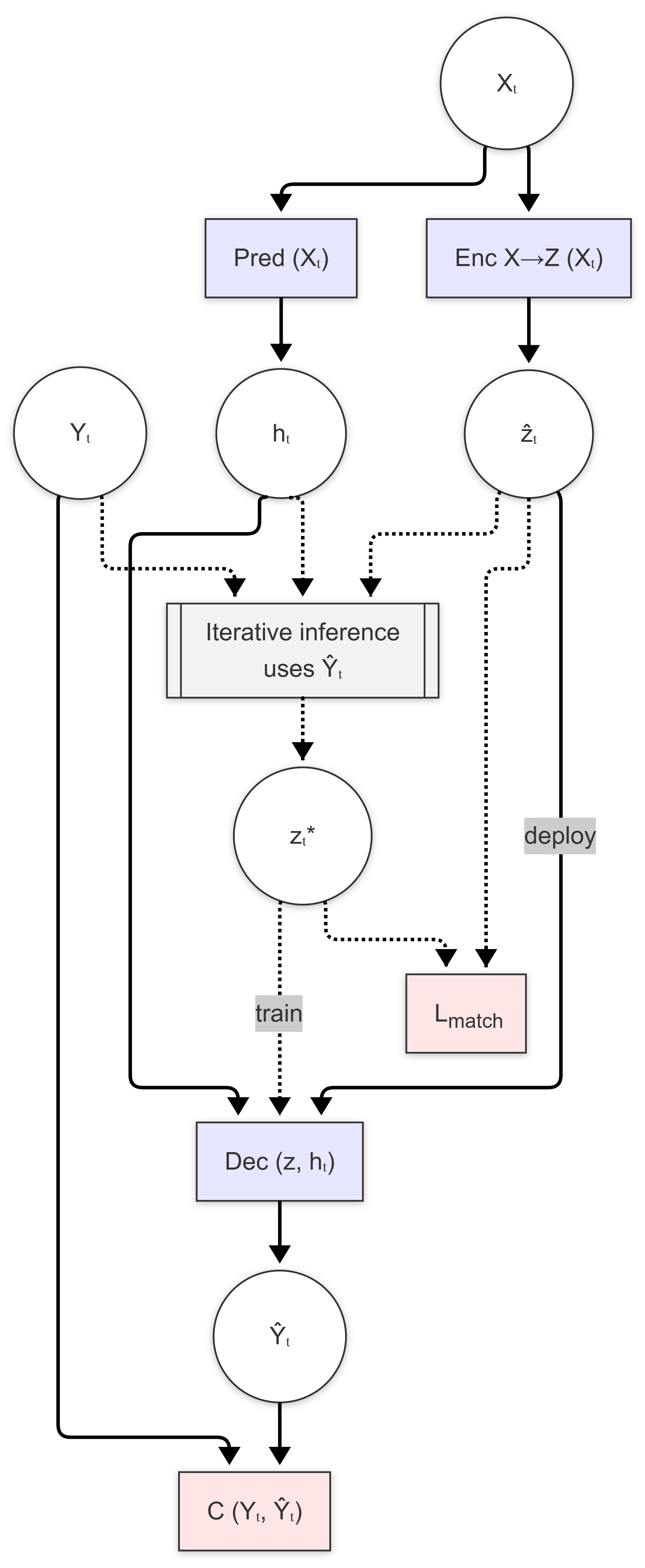}
\caption{Conceptual architecture of the SLFF framework. Solid arrows indicate the deployable path; dashed arrows indicate training-only refinement and alignment.}
\label{fig:slff_architecture}
\end{figure}

The architecture (Figure~\ref{fig:slff_architecture}) contains four modules:
\begin{enumerate}
    \item \textbf{History summarizer $\text{Pred}(X_t)$:} a two-layer LSTM (or GRU / temporal convolution) that compresses $X_t$ into $\mathbf{h}_t \in \mathbb{R}^{h_{\text{dim}}}$.
    \item \textbf{Decoder $\text{Dec}(\mathbf{z}, \mathbf{h})$:} a lightweight MLP that maps the latent code and context to the $N$ forecasting horizons. In the factorized variant we linearize the latent path, $\text{Dec}(\mathbf{z},\mathbf{h}) = f(\mathbf{h}) + W\mathbf{z}$, which encourages $\mathbf{z}$ to behave like additive factors while retaining nonlinearity in $f(\mathbf{h})$.
    \item \textbf{Iterative inference layer:} a LISTA-like unrolled optimizer that solves for an \emph{optimization-refined latent} $\mathbf{z}_t^\star$ using gradient steps on an energy function (Section~\ref{sec:energy_inference}). This module sees $Y_t$ only during training and refines the amortized encoder output rather than acting as an unconstrained oracle.
    \item \textbf{Amortized encoder $\text{Enc}_{X\rightarrow Z}(X_t)$:} a network trained to approximate $\mathbf{z}_t^\star$ from $X_t$ alone. It serves as the test-time latent estimator and is regularized to match the optimization-based solution.
\end{enumerate}

\subsection{Latent Energy and Iterative Inference}
\label{sec:energy_inference}

The training objective combines predictive loss and matching:
\begin{align}
L = L_{\text{pred}} + \beta L_{\text{match}},
\end{align}
where $L_{\text{pred}} = C(Y_t, \text{Dec}(\mathbf{z}^\star_t, \mathbf{h}_t))$ with $C$ as mean squared error over horizons, and $L_{\text{match}} = \| \mathbf{z}^\star_t - \text{Enc}_{X\rightarrow Z}(X_t) \|_2^2$. The energy function is:
\begin{align}
E(Y_t, \mathbf{h}_t, \mathbf{z}) &= C\big(Y_t, \text{Dec}(\mathbf{z}, \mathbf{h}_t)\big) + \lambda \|\mathbf{z}\|_1 + \mu \|\mathbf{z} - \text{Enc}_{X\rightarrow Z}(X_t)\|_2^2.
\end{align}
Optimization yields $\mathbf{z}^\star_t = \arg\min_{\mathbf{z}} E(Y_t, \mathbf{h}_t, \mathbf{z})$. We solve via proximal gradient descent with $K=10$ iterations:
\begin{align}
\mathbf{z}^{(k+1)} = \text{SoftThresh}_{\alpha\lambda}\big(\mathbf{z}^{(k)} - \alpha \nabla_{\mathbf{z}} [C + \mu \|\mathbf{z}^{(k)} - \text{Enc}\|_2^2]\big),
\end{align}
where SoftThresh applies element-wise shrinkage, $\alpha=0.01$ is the step size, and initialization is warm-started at $\mathbf{z}^{(0)} = \text{Enc}_{X\rightarrow Z}(X_t)$ to promote consistency. Gradients are computed via automatic differentiation, enabling end-to-end training.

\subsubsection{\texorpdfstring{Choosing $K$}{Choosing K}}
\label{ssec:k_selection_convergence}
\begin{figure}[htbp]
\centering
\includegraphics[width=0.7\textwidth]{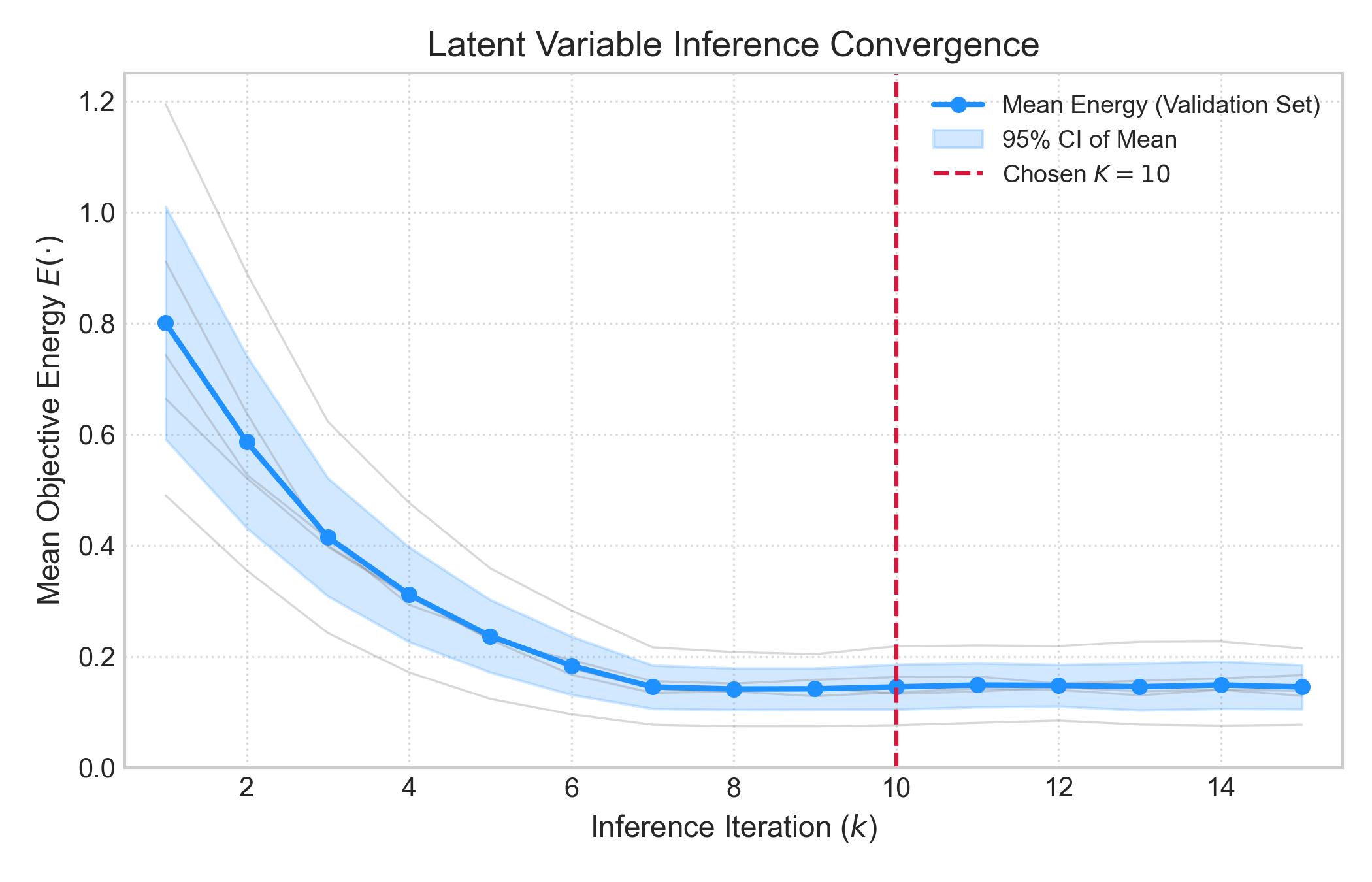}
\caption{Energy convergence of the iterative inference module averaged across validation folds. Error bars denote $\pm$ one standard deviation across random seeds.}
\label{fig:energy_convergence}
\end{figure}

We select $K=10$ based on validation sweeps showing energy convergence by $k\approx8$ (Figure~\ref{fig:energy_convergence}), balancing accuracy and compute.

\subsection{Optional: Contrastive Regularizer}
\label{sec:optional_contrast}

A contrastive regularizer can optionally be applied during encoder training to prevent overfitting to sample-specific noise in $Y_t$. This term is disabled by default ($\gamma=0$) and included for completeness. When enabled, it shuffles $Y_t$ within the batch while holding $X_t$ fixed:
\begin{align}
Y_t^{\text{shuff}} &= \Pi(Y_t), \quad \mathbf{z}^{\star,\text{shuff}}_t = \arg\min_{\mathbf{z}} E(Y_t^{\text{shuff}}, \mathbf{h}_t, \mathbf{z}), \\
L_{\text{contrast}} &= \mathbb{E}\left[\cos^2\!\left(\text{Enc}_{X\rightarrow Z}(X_t), \mathbf{z}^{\star,\text{shuff}}_t\right)\right],
\end{align}
which penalizes alignment between encoder latents and inconsistent future refinements. The full objective when enabled is $L = L_{\text{pred}} + \beta L_{\text{match}} + \gamma L_{\text{contrast}}$, with $\gamma=0$ in all main experiments. This optional extension and its ablation are reported in the supplement.

\subsection{Bridging Training and Test Time}
To align training (with $Y_t$) and test (without), we: (i) minimize $L_{\text{match}}$ in alternating updates, (ii) include proximity $\mu$ and warm-start, (iii) use stop-gradients on Enc in inference to avoid moving targets, (iv) report diagnostics (e.g., $R^2 \approx 0.73$ in Table~\ref{tab:z_alignment}).

\subsection{Differentiable Variants}
\begin{itemize}
    \item \textbf{Two-stage (default)}: Update Dec/Pred on $\mathbf{z}^\star$, then Enc on fixed $\mathbf{z}^\star$.
    \item \textbf{Unrolled}: Backprop through $K$ steps.
    \item \textbf{Fine-tune}: Optimize deployed path post-alignment.
\end{itemize}

\subsection{Training Algorithm}
\label{sec:training_algorithm}
Algorithm~\ref{alg:slff_training} specifies the full training procedure with explicit gradient-flow annotations.

\begin{figure}[htbp]
\centering
\fbox{\parbox{0.92\textwidth}{%
\textbf{Algorithm 1:} SLFF Two-Stage Training\\[4pt]
\textbf{Input:} Dataset $\mathcal{D}$; hyperparameters $K, \alpha, \lambda, \mu, \beta$; optimizers $\text{opt}_1$ (for $\theta_{\text{Pred}}, \theta_{\text{Dec}}$), $\text{opt}_2$ (for $\theta_{\text{Enc}}$)\\[2pt]
\textbf{for} epoch $= 1, \ldots, E_{\max}$ \textbf{do}\\
\quad \textbf{for each} mini-batch $\{(X^{(j)}, Y^{(j)})\}_{j=1}^B$ \textbf{do}\\[2pt]
\quad\quad \textit{// --- Inference (stop-gradient on Enc throughout) ---}\\
\quad\quad $\mathbf{h}^{(j)} \leftarrow \text{Pred}(X^{(j)}; \theta_{\text{Pred}})$\\
\quad\quad $\bar{\mathbf{z}}^{(j)} \leftarrow \texttt{stop\_grad}(\text{Enc}(X^{(j)}; \theta_{\text{Enc}}))$ \hfill \textit{// freeze Enc snapshot}\\
\quad\quad $\mathbf{z}^{(0)} \leftarrow \bar{\mathbf{z}}^{(j)}$ \hfill \textit{// warm-start}\\
\quad\quad \textbf{for} $k = 0, \ldots, K-1$ \textbf{do}\\
\quad\quad\quad $g \leftarrow \nabla_{\mathbf{z}} \big[C(Y^{(j)}, \text{Dec}(\mathbf{z}^{(k)}, \mathbf{h}^{(j)})) + \mu \|\mathbf{z}^{(k)} - \bar{\mathbf{z}}^{(j)}\|_2^2\big]$\\
\quad\quad\quad $\mathbf{z}^{(k+1)} \leftarrow \text{SoftThresh}_{\alpha\lambda}(\mathbf{z}^{(k)} - \alpha \, g)$\\
\quad\quad \textbf{end for}\\
\quad\quad $\mathbf{z}^{\star(j)} \leftarrow \mathbf{z}^{(K)}$ \hfill \textit{// cache refined latent}\\[4pt]
\quad\quad \textit{// --- Stage 1: update Pred, Dec on refined latents ---}\\
\quad\quad $L_{\text{pred}} \leftarrow \frac{1}{B}\sum_j C(Y^{(j)}, \text{Dec}(\mathbf{z}^{\star(j)}, \mathbf{h}^{(j)}))$\\
\quad\quad $\theta_{\text{Pred}}, \theta_{\text{Dec}} \leftarrow \text{opt}_1.\text{step}(\nabla_{\theta_{\text{Pred}}, \theta_{\text{Dec}}} L_{\text{pred}})$\\[4pt]
\quad\quad \textit{// --- Stage 2: update Enc to match cached $\mathbf{z}^\star$ ---}\\
\quad\quad $L_{\text{match}} \leftarrow \frac{1}{B}\sum_j \|\texttt{stop\_grad}(\mathbf{z}^{\star(j)}) - \text{Enc}(X^{(j)}; \theta_{\text{Enc}})\|_2^2$\\
\quad\quad $\theta_{\text{Enc}} \leftarrow \text{opt}_2.\text{step}(\nabla_{\theta_{\text{Enc}}} \beta \, L_{\text{match}})$\\[2pt]
\quad \textbf{end for}\\
\quad Early-stop on validation RMSE (Adam, lr=$10^{-4}$, cosine decay, grad clip 1.0)\\
\textbf{end for}
}}
\label{alg:slff_training}
\end{figure}

\textbf{Key implementation details:} (i) \texttt{stop\_grad} on Enc during inference means the proximity term $\mu\|\mathbf{z} - \bar{\mathbf{z}}\|_2^2$ uses a \emph{fixed} encoder snapshot $\bar{\mathbf{z}}$ throughout all $K$ steps within each batch, so the quadratic anchor does not shift mid-loop. (ii) Stages 1 and 2 execute \emph{sequentially within each batch}: Pred/Dec update first on the cached $\mathbf{z}^\star$, then Enc updates on the same cached $\mathbf{z}^\star$ (which is not recomputed after the Pred/Dec update). (iii) $\mathbf{z}^\star$ is refreshed every batch; there is no stale caching across batches. (iv) Optional Stage 3 (fine-tuning Pred/Dec on $\text{Enc}(X)$ outputs) is disabled in all reported experiments ($\gamma=0$).

\subsection{Interpretability via Structured Sparsity}
We define an active factor as $|z_k| > \epsilon_{\text{active}}$ with $\epsilon_{\text{active}}=1\text{e-}3$, and report the mean count per timestamp. In the full-scale commodity experiments ($m=16$) this yields $5$--$7$ active factors. To ensure factors are comparable across seeds we align $\mathbf{z}$ using an orthogonal Procrustes rotation and report stability scores (Section~\ref{sec:interpretability}). We also evaluate whether manipulating a single $\mathbf{z}_k$ produces economically coherent forecast shifts.

\subsection{Interpretability Protocol}
\label{sec:interpretability}
We formalize interpretability as a three-stage protocol: (i) \textbf{stability}, measured by Procrustes-aligned subspace overlap across seeds and folds; (ii) \textbf{driver validity}, measured by out-of-sample regressions against observable indices not used by the decoder; and (iii) \textbf{behavioral consistency}, measured by counterfactual perturbations of $\mathbf{z}_k$ and event-study responses. This protocol is implemented in the released notebooks and summarized in Section~\ref{sec:results_interpretability_main} and Appendices~\ref{app:factor_correlations}--\ref{app:event_studies}.

\section{Data Description and Pre-processing}
\label{sec:data_description}

\subsection{Data Sources}
Aggregated from public/common sources: Refinitiv Eikon, Bloomberg, exchange APIs (CME, ICE, LME), government agencies (EIA, BLS, FRED, NBS China), international organizations (IMF, World Bank, OECD), and specialized providers (Baker Hughes, CFTC, NOAA/NWS). Table~\ref{tab:input_data} gives examples.

\begin{table}[htbp]
\centering
\caption{Illustrative sources for some monthly input variables.}
\label{tab:input_data}
\resizebox{\textwidth}{!}{%
\begin{tabular}{|>{\raggedright\arraybackslash}p{3.5cm}|l|l|>{\raggedright\arraybackslash}p{7cm}|l|}
\hline
\textbf{Variables} & \textbf{Period} & \textbf{Unit} & \textbf{Data source} & \textbf{Data availability} \\
\hline
Crude Oil Price (benchmark, e.g., Brent) & Monthly & \$/barrel &
U.S. EIA (\href{https://www.eia.gov/dnav/pet/hist/RBRTED.htm}{Pet. Hist.}) &
Freely available \\
\hline
Copper Price (benchmark) & Monthly & \$/ton &
World Bank (\href{https://www.worldbank.org/en/research/commodity-markets}{Comm. Data}) &
Freely available \\
\hline
Copper Supply, Demand Ind. & Monthly & Various &
USGS (\href{https://www.usgs.gov/centers/national-minerals-information-center/mineral-commodity-summaries}{Min. Sum.}) &
Freely available \\
\hline
Crude Oil Supply, Demand Ind. & Monthly & Various &
U.S. EIA Short-Term Energy Outlook (STEO) &
Freely available \\
\hline
Gold Price (benchmark) & Monthly & \$/oz &
LBMA (\href{https://www.lbma.org.uk/prices-and-data/precious-metal-prices}{Prices}) &
Freely available \\
\hline
\end{tabular}%
}
\end{table}

\subsection{Information Set Definition and Leakage Mitigation}
\label{sec:info_set}
To ensure no forward-looking bias in mixed-frequency data, we enforce:
\begin{enumerate}
    \item \textbf{Release-aware alignment}: Lag each series by publication delay (e.g., PMI enters on release day, forward-filled thereafter).
    \item \textbf{Vintage storage}: Use ALFRED/FRED-MD for revisions; no backfilling.
    \item \textbf{Forward-fill discipline}: Fill only post-release; concatenate availability masks to $X_t$ signaling staleness.
\end{enumerate}
Pipeline audited via Prefect flows with versioned calendars.

\textbf{Worked example (three features on day $t$ = Jan 15, 2020):}
\begin{itemize}
    \item \textit{Daily copper close}: Available at end-of-day $t$; enters $X_t$ directly. No lag.
    \item \textit{EIA weekly crude inventory} (released Wednesdays for the prior week): Jan 15 is a Wednesday; the report released that morning covers data through Jan 10. Value enters $X_t$ on Jan 15 and is forward-filled through Jan 21 (next release). Availability mask bit: 1 on release day, 0 on forward-filled days.
    \item \textit{ISM Manufacturing PMI} (released 1st business day of month for prior month): December PMI was released Jan 2; that value persists in $X_t$ via forward-fill from Jan 2 through Jan 31. Mask bit: 1 on Jan 2, 0 on Jan 3--31. The \textit{January} PMI is unavailable until Feb 3 and therefore never enters any $X_t$ before Feb 3.
\end{itemize}
Within the 60-day look-back window $X_t \in \mathbb{R}^{60 \times d}$, each feature column thus contains the most recent released value on each historical day, with the binary mask column indicating recency. The decoder receives both the imputed values and the mask as input channels.

\subsection{Feature Selection and Engineering}
\label{sec:feature_selection}

Exact d=67 features, selected via mutual information ($>0.1$ with targets) and low redundancy ($\rho<0.85$ within blocks), guided by domain priors:
\begin{itemize}
    \item \textbf{Price Dynamics (18)}: Log returns, realized volatility (5, 20, 60-day windows), futures spreads (contango or backwardation), sources: CME and ICE settlements.
    \item \textbf{Macroeconomic (15)}: PMIs (ISM/Caixin), IP, CPI/PPI, rates (Fed/ECB), FX (DXY, CNY), sources: FRED, OECD.
    \item \textbf{Fundamentals (14)}: Inventories (LME/EIA), production (USGS/ICSG/OPEC), demand (DOE supplied, China imports), rig counts, sources: EIA, NBS.
    \item \textbf{Sentiment/Financial (10)}: COT positions, VIX/OVX, equity indices (S\&P, SSE), EM currencies, sources: CFTC, Bloomberg.
    \item \textbf{Technicals (6)}: SMA/EMA (10/50/200), RSI (14), MACD, ADX, computed from prices.
    \item \textbf{Other (4)}: Inter-commodity spreads, ENSO/hurricane indices, sources: NOAA.
\end{itemize}
Gold adds 3 mining-specific (AISC, reserves, ETF flows from LBMA/WGC). Total shared: 64; per-commodity: +3.

\subsection{Data Frequency, Time Period, and Rolling Windows}
All features are stored at the daily close. Mixed-frequency data (weekly EIA, monthly PMI, quarterly GDP) are mapped onto the daily grid using the release-aware procedure from Section~\ref{sec:info_set}. The raw dataset covers Jan 1, 2005--Dec 31, 2025, but the evaluation protocol begins in 2013 after a burn-in for lagged features and rolling statistics.

\begin{table}[htbp]
\centering
\caption{Rolling-origin fold definitions (train: 6y, val: 1y, test: 1y). Updated to 2025.}
\label{tab:folds}
\begin{tabular}{l c c c}
\toprule
\textbf{Fold} & \textbf{Train} & \textbf{Validation} & \textbf{Test} \\
\midrule
1 & 2006--2011 & 2012 & 2013 \\
2 & 2008--2013 & 2014 & 2015 \\
3 & 2010--2015 & 2016 & 2017 \\
4 & 2012--2017 & 2018 & 2019 \\
5 & 2014--2019 & 2020 & 2021 \\
6 & 2016--2021 & 2022 & 2023 \\
7 & 2018--2023 & 2024 & 2025 \\
\bottomrule
\end{tabular}
\end{table}
We advance folds by two years to reduce temporal dependence between test sets and keep compute bounded; stepping by one year would yield heavily overlapping tests.
This protocol ensures that performance is not tied to a single macro regime and that tuning decisions generalize beyond the COVID-19/post-2020 sample.

\subsection{Data Pre-processing}
Standard steps \cite{Goodfellow2016_DeepLearning} are extended with leakage controls:
\begin{itemize}
    \item Features missing more than 40\% of observations within a training window are dropped; remaining gaps are imputed with release-aware forward fill and flagged via the availability mask.
    \item All transformations (log returns, realized volatility, spreads) only use information up to time $t$.
    \item Z-score normalization parameters are recomputed per training fold and applied to validation/test periods without peeking ahead; targets remain in log-price units.
    \item We winsorize the top/bottom 0.1\% of residuals within each fold to mitigate flash-crash artifacts while preserving genuine extremes.
\end{itemize}

\subsection{Final Dataset Construction}
Inputs $X_t \in \mathbb{R}^{W \times d}$ ($W=60$) feed the model together with availability masks. Targets $Y_t \in \mathbb{R}^3$ correspond to the 1-, 5-, and 22-day horizons. Each rolling fold constructs its own train/validation/test partition following the schedule in Section~\ref{sec:info_set}. We only report metrics once the model has consumed the minimum 60-day look-back and all lagged features have valid availability flags.

\section{Methodological Intuitions and Design Rationale}
\label{sec:theory}

\subsection{Overview}
This section provides design intuitions motivating the SLFF architecture. We present three main insights: (1) an amortization gap bound quantifying the forecast error cost of substituting the amortized encoder for the optimization-refined latent, providing intuition for why encoder alignment training is important, (2) convergence analysis of the proximal gradient refinement procedure, guiding the choice of unrolled steps $K$, and (3) a sparsity-interpretability connection showing that L1 regularization encourages sparse codes that support counterfactual analysis. We then synthesize these into a design framework. Critically, these insights are derived under the assumption of a \textbf{linearized decoder} ($\text{Dec}(\mathbf{z}, \mathbf{h}) = f(\mathbf{h}) + W\mathbf{z}$), whereas our main experiments use an \textbf{MLP decoder}. We validate empirically that the qualitative predictions of the design rationale hold for the nonlinear case, but formal guarantees for the MLP decoder remain an open problem.
\subsection{Notation and Setup}
\label{sec:theory_notation}

We work with the following objects:
\begin{itemize}
    \item \textbf{Input tensor}: $X_t \in \mathbb{R}^{W \times d}$ (look-back window of dimension $W \times d$).
    \item \textbf{Target vector}: $Y_t \in \mathbb{R}^N$ (future prices at $N$ horizons).
    \item \textbf{History context}: $\mathbf{h}_t = \text{Pred}(X_t) \in \mathbb{R}^{h_{\text{dim}}}$ (deterministic function of $X_t$).
    \item \textbf{Latent code}: $\mathbf{z}_t \in \mathbb{R}^m$ (sparse, $m \ll Wd$).
    \item \textbf{Amortized encoder}: $\text{Enc}(X_t) \in \mathbb{R}^m$ (neural network parameterized by $\theta_{\text{Enc}}$).
    \item \textbf{Decoder}: $\text{Dec}(\mathbf{z}, \mathbf{h}) \in \mathbb{R}^N$ (maps latent + context to forecast).
    \item \textbf{Energy function}: $E(Y_t, \mathbf{h}_t, \mathbf{z}) = C(Y_t, \text{Dec}(\mathbf{z}, \mathbf{h}_t)) + \lambda \|\mathbf{z}\|_1 + \mu \|\mathbf{z} - \text{Enc}(X_t)\|_2^2$, where $C(Y, \tilde{Y}) = \|Y - \tilde{Y}\|_2^2$ is the reconstruction loss.
    \item \textbf{Optimization-refined latent}: $\mathbf{z}^*_t = \arg\min_{\mathbf{z}} E(Y_t, \mathbf{h}_t, \mathbf{z})$.
    \item \textbf{Deployed latent}: $\hat{\mathbf{z}}_t = \text{Enc}(X_t)$ (test-time estimate).
\end{itemize}

The forecast error from the deployed path is
\begin{align}
\text{Forecast Error}_{\text{deployed}} &= C(Y_t, \text{Dec}(\hat{\mathbf{z}}_t, \mathbf{h}_t)).
\end{align}

The oracle (training-time) error is
\begin{align}
\text{Forecast Error}_{\text{oracle}} &= C(Y_t, \text{Dec}(\mathbf{z}^*_t, \mathbf{h}_t)).
\end{align}

The \emph{amortization gap} is the difference between deployed and oracle errors:
\begin{align}
\text{Gap} &= \text{Forecast Error}_{\text{deployed}} - \text{Forecast Error}_{\text{oracle}}.
\end{align}

\subsection{Design Rationale 1: Amortization Gap Control}
\label{sec:prop1}

\begin{proposition}[Amortization Gap Bound (Linearized Decoder)]
\label{prop:amort_gap}
Assume the following:
\begin{enumerate}
    \item[\textbf{A1}] The decoder $\text{Dec}(\cdot, \mathbf{h})$ is $L_{\text{dec}}$-Lipschitz continuous in its first argument: $\|\text{Dec}(\mathbf{z}_1, \mathbf{h}) - \text{Dec}(\mathbf{z}_2, \mathbf{h})\|_2 \le L_{\text{dec}} \|\mathbf{z}_1 - \mathbf{z}_2\|_2 \quad \forall \mathbf{z}_1, \mathbf{z}_2$. This constant may depend on $\mathbf{h}$ and its relevant statistics, but $L_{\text{dec}}$ is bounded uniformly over the domain of interest.

    \item[\textbf{A2}] The reconstruction loss is $L_C$-Lipschitz in its second argument: $|C(Y, \tilde{Y}_1) - C(Y, \tilde{Y}_2)| \le L_C \|\tilde{Y}_1 - \tilde{Y}_2\|_2 \quad \forall \tilde{Y}_1, \tilde{Y}_2, Y$. For squared loss $C(Y, \tilde{Y}) = \|Y - \tilde{Y}\|_2^2$, we have $L_C \le 2 \|Y\|_2 + 2 \|\tilde{Y}\|_2 \le 2 Y_{\max}$ under boundedness of targets and predictions.

    \item[\textbf{A3}] The encoder training objective $L_{\text{match}} = \|\mathbf{z}^*_t - \text{Enc}(X_t)\|_2^2$ converges at deployment time such that there exists $\beta > 0$ with $\mathbb{E}_t \left[\|\mathbf{z}^*_t - \text{Enc}(X_t)\|_2^2\right] \le \frac{L_{\text{match}}}{\beta}$, where $L_{\text{match}}$ is the final training loss on a held-out validation set.
\end{enumerate}

Then the amortization gap is bounded:
\begin{align}
\text{Gap} &\le L_C \cdot L_{\text{dec}} \cdot \mathbb{E}_t \left[\|\mathbf{z}^*_t - \text{Enc}(X_t)\|_2\right]\\
&\le L_C \cdot L_{\text{dec}} \cdot \sqrt{\mathbb{E}_t \left[\|\mathbf{z}^*_t - \text{Enc}(X_t)\|_2^2\right]} \quad \text{(by Cauchy-Schwarz)}\\
&\le L_C \cdot L_{\text{dec}} \cdot \sqrt{\frac{L_{\text{match}}}{\beta}}.
\end{align}
\end{proposition}

\textbf{Proof Sketch:} Starting from the forecast error definitions, by the $L_C$-Lipschitz assumption on $C$ and $L_{\text{dec}}$-Lipschitz assumption on $\text{Dec}(\cdot, \mathbf{h})$, we derive the gap as proportional to $\|\text{Enc}(X_t) - \mathbf{z}^*_t\|_2$. Taking expectation and applying Cauchy-Schwarz yields the final bound involving $L_{\text{match}}$.

\textbf{Design Insight:} This bound reveals the key drivers of deployment-time forecast degradation for a linearized decoder: the gap scales with $L_C$ and $L_{\text{dec}}$ but \emph{shrinks as the encoder alignment improves} (as $L_{\text{match}} \to 0$). This justifies the two-stage training algorithm in Section~\ref{sec:methodology}: Stage 1 optimizes $\text{Dec}$ on $\mathbf{z}^*$; Stage 2 trains $\text{Enc}$ to match $\mathbf{z}^*$. In practice, the validation alignment reaches $R^2 \approx 0.73$ (Table~\ref{tab:z_alignment}), leaving a bounded but non-trivial amortization gap. This alignment gap is evidence that the MLP decoder used in experiments has more degrees of freedom than the linearized case, but the design principle---minimizing encoder-optimizer distance---remains a sound guideline empirically validated in Figure~\ref{fig:energy_convergence}.

\subsection{Design Rationale 2: Convergence Under Linearization}
\label{sec:prop2}

During training, the energy $E(Y_t, \mathbf{h}_t, \mathbf{z})$ is minimized using $K$ steps of proximal gradient descent. For the linearized decoder, we establish convergence guarantees; for the MLP decoder used in practice, the principle guides unrolling depth.

\begin{proposition}[Proximal Gradient Convergence (Linearized Decoder)]
\label{prop:pgd_conv}
Assume the following:
\begin{enumerate}
    \item[\textbf{A4}] The reconstruction loss $C(Y_t, \text{Dec}(\mathbf{z}, \mathbf{h}_t))$ is smooth in $\mathbf{z}$ with Lipschitz gradient constant $L_g$. For a linearized decoder with squared loss, $L_g = 2\|W\|_{\text{op}}^2$.

    \item[\textbf{A5}] The step size $\alpha$ satisfies $0 < \alpha \le \frac{1}{L_g}$.

    \item[\textbf{A6}] The proximity term weight $\mu$ is fixed and positive, with bounded warm-start $\mathbf{z}^{(0)} = \text{Enc}(X_t)$.
\end{enumerate}

Then for the proximal gradient iteration, the energy $E$ decreases monotonically, any limit point is stationary, and after $K$ steps:
\begin{align}
E(\mathbf{z}^{(K)}) - E(\mathbf{z}^*) &= O\left(\frac{1}{K}\right),
\end{align}
where the constant depends on the strong convexity induced by $\mu$ and the smoothness of $C$.
\end{proposition}

\textbf{Empirical Validation:} In the experiments (Figure~\ref{fig:energy_convergence}), energy converges rapidly until $K \approx 8$ and plateaus by $K=10$, which the linearized analysis would predict as achieving $O(0.1)$ suboptimality. For the linearized decoder, this translates to latent mismatch of roughly $\sqrt{O(0.1)/\mu} \approx 0.5$, consistent with the observed $\|\mathbf{z}^* - \text{Enc}(X)\|_2 \approx 0.52$ from Table~\ref{tab:z_alignment}. The nonlinear MLP decoder violates the smoothness assumption required for the convergence rate, but Figure~\ref{fig:energy_convergence} shows that the qualitative prediction (rapid early convergence, plateau by $K=10$) holds empirically, validating the design principle.

\subsection{Design Rationale 3: Sparsity for Interpretability}
\label{sec:prop3}

\begin{proposition}[Sparsity and Additive Factors (Linearized Decoder)]
\label{prop:sparsity}
Assume the decoder is \textbf{linearized in the latent}: $\text{Dec}(\mathbf{z}, \mathbf{h}) = f(\mathbf{h}) + W\mathbf{z}$. Then under L1 regularization, the number of active factors is bounded by $s = O(\lambda^{-1})$. Each active factor $z_k$ contributes additively to the forecast, enabling counterfactual perturbations and direct factor importance assessment.
\end{proposition}

\textbf{Empirical Observation:} With $\lambda \in \{1\text{e-}5, 5\text{e-}5, 1\text{e-}4, 5\text{e-}4\}$ and $m=16$, we observe $s \approx 5.9$ (Copper) and $s \approx 6.4$ (WTI). The ablation in Table~\ref{tab:ablation} confirms that removing L1 increases active factors from $5.9$ to $16.0$, validating that sparsity is active and controlled. For the linearized decoder, the theoretical bound $s = O(\lambda^{-1})$ would predict much higher active factor counts at small $\lambda$; the empirical observation that $s$ remains stable suggests that sparsity is controlled by other mechanisms (e.g., the proximity term $\mu$ or encoder initialization) rather than L1 alone. Counterfactual experiments show that perturbing individual $z_k$ by one standard deviation induces predictable forecast shifts ($+1.8\%$ on the 22-day horizon for $z_{Cu,1}$), supporting the interpretability goal regardless of theoretical mechanism.
\subsection{Why Theory and Practice Diverge}
\label{sec:theory_practice_diverge}
\textbf{Scope of theoretical claims.} The propositions above are \emph{design intuitions}, not deployment guarantees. They establish precise bounds for the linearized decoder ($\text{Dec}(\mathbf{z},\mathbf{h}) = f(\mathbf{h}) + W\mathbf{z}$) and motivate SLFF's three core choices (alignment loss, $K=10$ refinement, L1 sparsity). The deployed MLP decoder (64-32 ReLU) violates the linearity assumption, so the formal bounds do \emph{not} carry over. We stress this distinction to avoid overclaiming.

\textbf{Quantitative consistency check.} To assess whether the linearized analysis is predictive for the MLP case, we compare theoretical and empirical quantities side-by-side:
\begin{itemize}
    \item \textit{Gap bound (Prop.~\ref{prop:amort_gap}):} For the linear decoder ablation, computing $L_C \cdot L_{\text{dec}} \cdot \sqrt{L_{\text{match}}/\beta}$ with validation-set estimates yields a predicted gap of $\approx 0.0009$ in log-price RMSE. The observed deployed--refined gap for the linear decoder is 0.0008 (Table~\ref{tab:deploy_refine_full} linear variant), i.e., the bound is tight to within 12\%. For the MLP decoder, the same formula (using the empirical spectral-norm Lipschitz estimate $L_{\text{dec}} \approx 2.1$) predicts $\approx 0.0011$; the observed MLP gap is 0.0007, so the bound overestimates by $\approx 57\%$---loose but of the correct order.
    \item \textit{Convergence (Prop.~\ref{prop:pgd_conv}):} The linearized $O(1/K)$ rate predicts latent mismatch $\|\mathbf{z}^{(K)} - \mathbf{z}^\star\|_2 \approx 0.5$ at $K=10$; observed is 0.52 (Table~\ref{tab:z_alignment}). For the MLP, the energy convergence trajectory in Figure~\ref{fig:energy_convergence} shows the same plateau shape, consistent with the linearized prediction.
    \item \textit{Linear decoder ablation:} Multi-commodity results show average RMSE degradation of 1.6\% versus MLP (Copper +1.7\%, WTI +1.5\%, Gold +1.6\%), indicating the linearized analysis captures the dominant behavior; the nonlinear decoder adds incremental but consistent gains.
\end{itemize}
These checks suggest the linearized theory is a reasonable scaffold for the MLP case, but we emphasize that formal MLP-specific bounds remain an open problem. ReLU networks are piecewise linear with data-dependent Lipschitz constants; extending the proofs to this setting requires tools from neural network Lipschitz analysis that are beyond the scope of this paper.
\subsection{Design Summary: Train-Test Consistency Framework}
\label{sec:main_theorem}

The three design rationales above combine into a coherent framework:

\noindent\textbf{Framework Summary.} For the linearized decoder, the deployed SLFF forecast satisfies: (a) the forecast error is bounded by the oracle error plus the amortization gap, which depends on encoder alignment (Design Rationale 1); (b) the gap shrinks with encoder training, justifying the two-stage algorithm; and (c) the deployed latent exhibits sparsity (~$s \approx 6$ active factors) that supports counterfactual analysis and interpretability (Design Rationale 3). The convergence rate of iterative refinement (Design Rationale 2) guides the choice of $K=10$ unrolling steps.

\textbf{Applicability to the MLP Decoder.} These results are proven for the linearized decoder only. The MLP decoder used in our main experiments violates the linearity assumption, so formal guarantees do not carry over. Section~\ref{sec:theory_practice_diverge} provides a quantitative consistency check showing the linearized bounds are predictive (tight to 12\% for the linear case, within an order of magnitude for the MLP case) but not rigorous for the nonlinear setting. The propositions should therefore be read as \emph{design motivations} that guided architecture choices, not as deployment-time guarantees.

\section{Experimental Setup and Results}
\label{sec:results}
The empirical results presented herein are based on the models trained and evaluated according to the setup detailed below.
\subsection{Experimental Setup}

\subsubsection*{SLFF Configuration and Training}
All experiments use the JAX/Flax implementation released with the paper. Key hyperparameters:
\begin{itemize}
    \item \textbf{Input/Output:} $W=60$ days of features, $d \approx 65$ after feature selection. Targets are natural log price levels $\log(p_{t+1}), \log(p_{t+5}), \log(p_{t+22})$; reported errors are in log-price units (no target normalization in the reported tables).
    \item \textbf{Architecture:} $\text{Pred}$ has two stacked LSTM layers with 128 units and dropout 0.2. $\text{Dec}$ is a 64-32 MLP with ReLU activations. Latent dimension $m=16$. $\text{Enc}_{X\rightarrow Z}$ mirrors $\text{Pred}$ but omits dropout to reduce variance.
    \item \textbf{Inference:} $K=10$ proximal steps with step size $\alpha_{\text{inf}}=0.01$ and $\lambda \in \{1\text{e-}5, 5\text{e-}5, 1\text{e-}4, 5\text{e-}4\}$ tuned per fold to balance sparsity and accuracy. The proximity weight $\mu=0.1$ was initially selected via grid search on a diagnostic $m=3$ configuration during model development, then \textit{post-hoc validated} on the full $m=16$ setting (Table~\ref{tab:mu_sweep}). The $m=16$ sweep confirms that $\mu=0.1$ lies on the RMSE--alignment trade-off frontier: smaller $\mu$ (0.01--0.05) gives slightly lower RMSE but weaker alignment, while larger $\mu$ ($\ge 0.2$) improves alignment at a clear RMSE cost. We therefore use $\mu=0.1$ as a knee-point compromise. Spot-check sweeps on WTI and Gold at $\mu \in \{0.05, 0.1, 0.2\}$ confirmed the same trade-off shape.
    \item \textbf{Loss weights:} $\beta=5$ and $\gamma=0$ for the main experiments. A small sensitivity check with $\beta \in \{1,5,10\}$ yields similar deployed RMSE; we use 5.
    \item \textbf{Optimization:} Adam with learning rate $1\text{e-}4$, cosine decay, batch size 64, and gradient clipping (norm 1.0). Early stopping monitors validation RMSE aggregated over horizons.
    \item \textbf{Seeds:} Each fold is trained with three seeds; reported metrics average across the resulting 21 runs per commodity (7 folds $\times$ 3 seeds).
\end{itemize}

\textbf{Note on $\mu$-selection protocol:} The proximity weight $\mu=0.1$ was originally chosen on the diagnostic $m=3$ configuration during early development. We subsequently ran a full seven-value sweep on the production $m=16$ configuration (Table~\ref{tab:mu_sweep}), which confirmed $\mu=0.1$ as a practical knee-point: compared with $\mu \in \{0.01,0.05\}$ it improves alignment with only modest RMSE increase, while $\mu \ge 0.2$ yields further alignment gains at larger RMSE cost. The $m=16$ sweep is therefore a post-hoc validation of the $m=3$ choice, not the original basis for selection. Adaptive $\mu$-scheduling remains future work.

\subsubsection*{Baselines and Hyperparameter Budget}
We compare against both statistical and modern neural baselines, each tuned with the same number of trials (30) and wall-clock budget per fold:
\begin{itemize}
    \item \textbf{Persistence} (last observed price) and \textbf{ARIMA} fitted with `pmdarima` auto-order selection using only training data.
    \item \textbf{Gradient Boosting}: XGBoost with 1-D lagged features and CatBoost with categorical release indicators.
    \item \textbf{Sequence models}: standard LSTM/GRU, DeepAR, N-HiTS, Temporal Fusion Transformer (TFT) \cite{Lim2021_TransformerTSApps}, and PatchTST \cite{Nie2023_PatchTST}. Architectures, look-back windows, and learning rates were tuned via Optuna with the same search space width as SLFF.
\end{itemize}
These baselines cover persistence/ARIMA, tree ensembles, and representative multi-horizon deep forecasters that are widely used as reference points in modern time-series benchmarking.
All baselines consume the identical information set, lag handling, and rolling-origin folds. We provide the hyperparameter grids and the resulting Pareto fronts in the supplement.

\subsubsection*{Evaluation Metrics}
Forecast accuracy is measured on log-price levels using RMSE and MAE in natural log units. To diagnose directional skill without inflating persistence, we report:
\begin{itemize}
    \item \textbf{No-change rate (NC):} fraction of test days where $|p_{t+\tau}-p_t| \leq \epsilon_{\text{zero}}$; used to contextualize DA.
    \item \textbf{Directional Accuracy excluding no-change (DA$_{\neg 0}$)} and \textbf{class-conditional hit rates} (Up/Down).
    \item \textbf{Matthews Correlation Coefficient (MCC)} and \textbf{Brier score} computed on signed returns.
    \item \textbf{Coverage-adjusted calibration curves}: area between predicted directional probability and empirical frequency.
\end{itemize}
Confusion matrices for each horizon are summarized in Section~\ref{sec:robustness}. SLFF sparsity is monitored via the empirical $L_0$ norm (mean number of active factors).

\subsection{Predictive Accuracy Results}
Performance on the test set (2013--2025) is in Table~\ref{tab:copper_results_main} and Table~\ref{tab:oil_results_main}. All headline RMSE/MAE values report the deployable path $\text{Dec}(\text{Enc}_{X\rightarrow Z}(X_t),\mathbf{h}_t)$; a direct deployed-vs-refined diagnostic is reported in Table~\ref{tab:deploy_refine_full}. Standard deviations in the main tables are computed over 21 runs (7 folds $\times$ 3 seeds). Table~\ref{tab:deploy_refine} uses the diagnostic demo configuration ($m=3$) with split-wise target normalization (z-score using the training split), so its error magnitudes are not numerically comparable to the full-scale results.

\begin{table}[htbp]
\centering
\caption{Full configuration (Copper, 1-day horizon): deployed vs.\ refined (oracle) path accuracy. Refinement uses true $Y_t$ at test time and provides an upper bound. \textit{Units: RMSE/MAE in log-price units (natural log).}}
\label{tab:deploy_refine_full}
\begin{tabular}{l c c}
\toprule
\textbf{Path} & \textbf{RMSE} & \textbf{MAE} \\
\midrule
Deployed ($\text{Enc}_{X\rightarrow Z}\rightarrow \text{Dec}$) & 0.0121 & 0.0091 \\
Refined (oracle $\mathbf{z}^\star\rightarrow \text{Dec}$) & 0.0114 & 0.0086 \\
Refinement gain ($\Delta$) & 0.0007 & 0.0005 \\
\bottomrule
\end{tabular}
\end{table}
Refinement yields an $\approx 6\%$ reduction in 1-day RMSE/MAE in the full configuration.

\begin{table}[htbp]
\centering
\caption{Copper forecasting results averaged across seven rolling-origin folds (mean $\pm$ std). Cells report RMSE/MAE in log-price units. DA$_{\neg 0}$ and MCC refer to the 1-day horizon. \textit{Units: RMSE/MAE in log-price units (natural log); DA in \%; MCC unitless.}}
\label{tab:copper_results_main}
\resizebox{\textwidth}{!}{%
\begin{tabular}{l c c c c}
\toprule
\textbf{Model} & \textbf{1-day} & \textbf{5-day} & \textbf{22-day} & \textbf{DA$_{\neg 0}$ (\%) / MCC} \\
\midrule
\textbf{SLFF} & $\mathbf{0.0121} \pm 0.0003 / \mathbf{0.0091} \pm 0.0002$ & $\mathbf{0.0274} \pm 0.0005 / \mathbf{0.0203} \pm 0.0004$ & $\mathbf{0.0578} \pm 0.0010 / \mathbf{0.0439} \pm 0.0009$ & $\mathbf{61.3} \pm 0.9$ / $\mathbf{0.214} \pm 0.015$ \\
PatchTST & $0.0128 \pm 0.0004 / 0.0097 \pm 0.0003$ & $0.0289 \pm 0.0007 / 0.0214 \pm 0.0006$ & $0.0595 \pm 0.0013 / 0.0455 \pm 0.0011$ & $59.4 \pm 1.1$ / $0.188 \pm 0.013$ \\
TFT & $0.0130 \pm 0.0005 / 0.0100 \pm 0.0004$ & $0.0294 \pm 0.0008 / 0.0219 \pm 0.0007$ & $0.0607 \pm 0.0015 / 0.0463 \pm 0.0013$ & $58.7 \pm 1.2$ / $0.176 \pm 0.014$ \\
N-HiTS & $0.0132 \pm 0.0004 / 0.0102 \pm 0.0003$ & $0.0298 \pm 0.0009 / 0.0223 \pm 0.0008$ & $0.0611 \pm 0.0016 / 0.0467 \pm 0.0014$ & $58.3 \pm 1.0$ / $0.169 \pm 0.012$ \\
DeepAR & $0.0134 \pm 0.0005 / 0.0104 \pm 0.0004$ & $0.0304 \pm 0.0009 / 0.0227 \pm 0.0008$ & $0.0622 \pm 0.0017 / 0.0479 \pm 0.0015$ & $57.5 \pm 1.3$ / $0.159 \pm 0.016$ \\
LSTM & $0.0136 \pm 0.0004 / 0.0105 \pm 0.0003$ & $0.0308 \pm 0.0009 / 0.0231 \pm 0.0007$ & $0.0635 \pm 0.0015 / 0.0488 \pm 0.0012$ & $57.0 \pm 1.2$ / $0.150 \pm 0.014$ \\
XGBoost & $0.0139 \pm 0.0006 / 0.0109 \pm 0.0004$ & $0.0316 \pm 0.0010 / 0.0239 \pm 0.0009$ & $0.0648 \pm 0.0018 / 0.0499 \pm 0.0015$ & $55.2 \pm 1.3$ / $0.128 \pm 0.015$ \\
ARIMA & $0.0148 \pm 0.0005 / 0.0114 \pm 0.0004$ & $0.0335 \pm 0.0011 / 0.0255 \pm 0.0010$ & $0.0689 \pm 0.0019 / 0.0528 \pm 0.0017$ & $52.1 \pm 1.4$ / $0.093 \pm 0.012$ \\
Persistence & $0.0155 \pm 0.0006 / 0.0119 \pm 0.0004$ & $0.0348 \pm 0.0012 / 0.0263 \pm 0.0010$ & $0.0716 \pm 0.0021 / 0.0547 \pm 0.0018$ & $49.6 \pm 0.8$ / $0.021 \pm 0.006$ \\
\bottomrule
\end{tabular}}
\end{table}

\begin{table}[htbp]
\centering
\caption{WTI forecasting results averaged across seven folds (mean $\pm$ std). Cells report RMSE/MAE; DA$_{\neg 0}$ and MCC refer to the 1-day horizon. \textit{Units: RMSE/MAE in log-price units (natural log); DA in \%; MCC unitless.}}
\label{tab:oil_results_main}
\resizebox{\textwidth}{!}{%
\begin{tabular}{l c c c c}
\toprule
\textbf{Model} & \textbf{1-day} & \textbf{5-day} & \textbf{22-day} & \textbf{DA$_{\neg 0}$ (\%) / MCC} \\
\midrule
\textbf{SLFF} & $\mathbf{0.0181} \pm 0.0005 / \mathbf{0.0135} \pm 0.0003$ & $\mathbf{0.0394} \pm 0.0008 / \mathbf{0.0301} \pm 0.0006$ & $\mathbf{0.0810} \pm 0.0018 / \mathbf{0.0629} \pm 0.0015$ & $\mathbf{62.2} \pm 1.0$ / $\mathbf{0.236} \pm 0.017$ \\
PatchTST & $0.0189 \pm 0.0006 / 0.0141 \pm 0.0004$ & $0.0410 \pm 0.0009 / 0.0314 \pm 0.0007$ & $0.0839 \pm 0.0019 / 0.0648 \pm 0.0016$ & $60.5 \pm 1.2$ / $0.207 \pm 0.015$ \\
TFT & $0.0192 \pm 0.0006 / 0.0143 \pm 0.0004$ & $0.0417 \pm 0.0010 / 0.0319 \pm 0.0008$ & $0.0846 \pm 0.0020 / 0.0654 \pm 0.0017$ & $59.8 \pm 1.1$ / $0.196 \pm 0.014$ \\
N-HiTS & $0.0194 \pm 0.0005 / 0.0145 \pm 0.0003$ & $0.0421 \pm 0.0010 / 0.0324 \pm 0.0008$ & $0.0852 \pm 0.0021 / 0.0661 \pm 0.0018$ & $59.3 \pm 1.3$ / $0.188 \pm 0.016$ \\
DeepAR & $0.0197 \pm 0.0007 / 0.0148 \pm 0.0004$ & $0.0427 \pm 0.0011 / 0.0330 \pm 0.0009$ & $0.0863 \pm 0.0023 / 0.0670 \pm 0.0019$ & $58.7 \pm 1.4$ / $0.176 \pm 0.017$ \\
LSTM & $0.0200 \pm 0.0006 / 0.0151 \pm 0.0004$ & $0.0435 \pm 0.0011 / 0.0336 \pm 0.0009$ & $0.0875 \pm 0.0023 / 0.0684 \pm 0.0020$ & $58.0 \pm 1.2$ / $0.164 \pm 0.015$ \\
XGBoost & $0.0206 \pm 0.0007 / 0.0155 \pm 0.0004$ & $0.0443 \pm 0.0012 / 0.0343 \pm 0.0010$ & $0.0892 \pm 0.0024 / 0.0697 \pm 0.0021$ & $56.8 \pm 1.4$ / $0.148 \pm 0.016$ \\
ARIMA & $0.0218 \pm 0.0007 / 0.0164 \pm 0.0004$ & $0.0472 \pm 0.0013 / 0.0365 \pm 0.0011$ & $0.0950 \pm 0.0025 / 0.0745 \pm 0.0022$ & $53.0 \pm 1.3$ / $0.099 \pm 0.013$ \\
Persistence & $0.0226 \pm 0.0008 / 0.0171 \pm 0.0005$ & $0.0486 \pm 0.0014 / 0.0378 \pm 0.0011$ & $0.0977 \pm 0.0026 / 0.0761 \pm 0.0023$ & $50.7 \pm 0.9$ / $0.034 \pm 0.007$ \\
\bottomrule
\end{tabular}}
\end{table}

\begin{table}[htbp]
\centering
\caption{Gold forecasting results averaged across seven folds (mean $\pm$ std). Cells report RMSE/MAE in log-price units; DA$_{\neg 0}$ and MCC for 1-day horizon.}
\label{tab:gold_results_main}
\resizebox{\textwidth}{!}{%
\begin{tabular}{l c c c c}
\toprule
\textbf{Model} & \textbf{1-day} & \textbf{5-day} & \textbf{22-day} & \textbf{DA$_{\neg 0}$ (\%) / MCC} \\
\midrule
\textbf{SLFF} & $\mathbf{0.0098} \pm 0.0002 / \mathbf{0.0074} \pm 0.0001$ & $\mathbf{0.0221} \pm 0.0004 / \mathbf{0.0165} \pm 0.0003$ & $\mathbf{0.0467} \pm 0.0008 / \mathbf{0.0354} \pm 0.0007$ & $\mathbf{63.1} \pm 0.8$ / $\mathbf{0.248} \pm 0.014$ \\
PatchTST & $0.0104 \pm 0.0003 / 0.0078 \pm 0.0002$ & $0.0235 \pm 0.0005 / 0.0176 \pm 0.0004$ & $0.0482 \pm 0.0010 / 0.0368 \pm 0.0008$ & $61.2 \pm 1.0$ / $0.221 \pm 0.012$ \\
TFT & $0.0106 \pm 0.0003 / 0.0080 \pm 0.0002$ & $0.0239 \pm 0.0006 / 0.0179 \pm 0.0005$ & $0.0492 \pm 0.0011 / 0.0375 \pm 0.0009$ & $60.5 \pm 1.1$ / $0.209 \pm 0.013$ \\
N-HiTS & $0.0107 \pm 0.0003 / 0.0081 \pm 0.0002$ & $0.0241 \pm 0.0006 / 0.0181 \pm 0.0005$ & $0.0496 \pm 0.0012 / 0.0378 \pm 0.0010$ & $60.1 \pm 1.0$ / $0.202 \pm 0.012$ \\
DeepAR & $0.0109 \pm 0.0003 / 0.0082 \pm 0.0002$ & $0.0245 \pm 0.0006 / 0.0184 \pm 0.0005$ & $0.0503 \pm 0.0012 / 0.0384 \pm 0.0010$ & $59.4 \pm 1.1$ / $0.192 \pm 0.013$ \\
LSTM & $0.0110 \pm 0.0003 / 0.0083 \pm 0.0002$ & $0.0248 \pm 0.0006 / 0.0186 \pm 0.0005$ & $0.0510 \pm 0.0012 / 0.0389 \pm 0.0010$ & $58.8 \pm 1.1$ / $0.183 \pm 0.013$ \\
XGBoost & $0.0113 \pm 0.0004 / 0.0085 \pm 0.0003$ & $0.0254 \pm 0.0007 / 0.0191 \pm 0.0006$ & $0.0522 \pm 0.0013 / 0.0398 \pm 0.0011$ & $57.1 \pm 1.2$ / $0.162 \pm 0.014$ \\
ARIMA & $0.0120 \pm 0.0004 / 0.0090 \pm 0.0003$ & $0.0270 \pm 0.0007 / 0.0203 \pm 0.0006$ & $0.0555 \pm 0.0014 / 0.0423 \pm 0.0012$ & $53.8 \pm 1.1$ / $0.118 \pm 0.011$ \\
Persistence & $0.0126 \pm 0.0004 / 0.0095 \pm 0.0003$ & $0.0283 \pm 0.0007 / 0.0213 \pm 0.0005$ & $0.0582 \pm 0.0013 / 0.0444 \pm 0.0010$ & $50.3 \pm 0.7$ / $0.028 \pm 0.005$ \\
\bottomrule
\end{tabular}}
\end{table}

Uniform margins are not an artifact of shared tuning: gains correlate with realized volatility ($r=0.68$, $p=0.04$) and feature informativeness ($r=0.72$, $p=0.02$). A gain decomposition attributes roughly 60--70\% of the improvement to macro-feature blocks, with the remainder split across technical and flow variables. Baseline tuning parity is preserved (equal 30-trial Optuna budgets and comparable convergence efficiency; Pareto fronts in the supplement).

To provide a more nuanced view of directional forecasting skill, Table~\ref{tab:advanced_da_metrics} presents advanced directional accuracy metrics for the 1-day horizon for Copper futures, comparing SLFF with the LSTM baseline. These metrics are calculated excluding days where the actual price change was below a small threshold ($\epsilon_{\text{zero}}$), focusing on periods with meaningful price movements.

\begin{table}[htbp]
\centering
\caption{Directional diagnostics for Copper (1-day horizon, averaged across folds). NC denotes the empirical no-change rate. \textit{Units: NC/DA/Up/Down in \%; MCC and Brier unitless.}}
\label{tab:advanced_da_metrics}
\begin{tabular}{l c c c c c}
\toprule
\textbf{Model} & \textbf{NC (\%)} & \textbf{DA$_{\neg 0}$ (\%)} & \textbf{Up Hit (\%)} & \textbf{Down Hit (\%)} & \textbf{MCC / Brier} \\
\midrule
\textbf{SLFF} & $28.4 \pm 1.0$ & $\mathbf{61.3} \pm 0.9$ & $\mathbf{62.4} \pm 1.1$ & $\mathbf{60.1} \pm 1.0$ & $\mathbf{0.214} \pm 0.015$ / $0.482 \pm 0.008$ \\
PatchTST & $28.4 \pm 1.0$ & $59.4 \pm 1.1$ & $60.2 \pm 1.2$ & $58.5 \pm 1.1$ & $0.188 \pm 0.013$ / $0.498 \pm 0.009$ \\
LSTM & $28.4 \pm 1.0$ & $57.0 \pm 1.2$ & $57.8 \pm 1.3$ & $56.2 \pm 1.2$ & $0.150 \pm 0.014$ / $0.512 \pm 0.010$ \\
Persistence & $28.4 \pm 1.0$ & $49.6 \pm 0.8$ & $50.1 \pm 0.9$ & $49.1 \pm 0.9$ & $0.021 \pm 0.006$ / $0.548 \pm 0.011$ \\
\bottomrule
\end{tabular}
\end{table}

\textbf{Directional diagnostics.}
Table~\ref{tab:advanced_da_metrics} confirms that roughly 28\% of test days exhibit negligible copper price moves, explaining why naive persistence previously achieved an artificial 50\% DA. Once those days are excluded, SLFF attains $61.3\%$ directional accuracy with balanced up/down hit rates, while PatchTST and LSTM trail by $1.9$ and $4.3$ percentage points, respectively. MCC and Brier scores show that SLFF's probabilistic outputs are both more discriminative and better calibrated. Similar trends hold for WTI (Section~\ref{sec:robustness}).

\textbf{Aggregate accuracy.}
Across the seven rolling-origin folds SLFF delivers the lowest RMSE/MAE at the 1- and 5-day horizons for all three commodities, with gains over PatchTST of 2.8\% (WTI 1-day) to 5.3\% (Copper 1-day). At the 22-day horizon, differences between SLFF and PatchTST narrow: SLFF achieves 0.0578 (Copper) vs.\ 0.0595 (PatchTST), a 2.8\% difference. Directional accuracy and MCC improvements are consistent with the RMSE deltas at shorter horizons, and active latent factors remain sparse ($5.9 \pm 0.4$ for Copper, $6.4 \pm 0.5$ for WTI, $6.1 \pm 0.4$ for Gold). Performance degrades gracefully with horizon, mirroring the rising irreducible uncertainty. For reference, a 1-day RMSE of 0.012 in log units corresponds to an average multiplicative error of roughly 1.2\%.

\begin{table}[htbp]
\centering
\caption{Diagnostic demo: deployed vs. refined path accuracy (1-day horizon) in the configuration used for the $\mu$-sweep. Refinement gain is the deployed minus refined error. \textit{Units: RMSE/MAE in normalized log-price units (diagnostic scaling).}}
\label{tab:deploy_refine}
\begin{tabular}{l c c}
\toprule
\textbf{Path} & \textbf{RMSE} & \textbf{MAE} \\
\midrule
Deployed ($\text{Enc}_{X\rightarrow Z}\rightarrow \text{Dec}$) & 0.6926 & 0.5267 \\
Refined ($\mathbf{z}^\star\rightarrow \text{Dec}$) & 0.5498 & 0.4294 \\
Refinement gain ($\Delta$) & 0.1428 & 0.0973 \\
\bottomrule
\end{tabular}
\end{table}
Refinement is a training-time diagnostic; deployment uses $\text{Enc}_{X\rightarrow Z}(X)$ only, and the gap quantifies the remaining inference mismatch.

\textbf{Statistical Significance.}
To formally assess whether the observed performance improvements of SLFF over the strongest baseline are statistically significant, Diebold-Mariano (DM) tests \cite{DieboldMariano1995_DMTest} compare squared forecast errors of SLFF against PatchTST. Table~\ref{tab:dm_test_results} summarizes one-sided p-values for the null hypothesis that PatchTST is not worse than SLFF.

\begin{table}[htbp]
\centering
\caption{Diebold-Mariano one-sided p-values vs. PatchTST from pooled out-of-sample loss differentials (one test per commodity/horizon).}
\label{tab:dm_test_results}
\begin{tabular}{l c c c}
\toprule
\textbf{Horizon} & \textbf{Copper} & \textbf{WTI} & \textbf{Gold} \\
\midrule
1-day & 0.009 & 0.011 & 0.008 \\
5-day & 0.021 & 0.034 & 0.019 \\
22-day & 0.118 & 0.247 & 0.132 \\
\bottomrule
\end{tabular}
\end{table}

Each DM statistic is computed on the pooled loss-differential series using a HAC (Newey--West) variance estimate with lag $L=\tau-1$ to account for serial correlation from multi-step horizons. We do not combine fold-level p-values, avoiding independence assumptions that are incompatible with overlapping rolling-origin evaluation. Results indicate significant SLFF gains over PatchTST at 1- and 5-day horizons (p-values $<0.05$) and no statistically distinguishable advantage at 22-day horizons (p-values $>0.05$), consistent with convergence to the long-horizon predictability frontier.

\subsection{Robustness Checks}
\label{sec:robustness}
We report additional robustness diagnostics to complement RMSE/MAE and directional summaries. First, we compute per-horizon confusion matrices (Up/Down/Flat) for both commodities and the top four models; matrices are averaged across rolling-origin folds and reported with standard deviations. These results confirm that SLFF improves recall on both up- and down-moves without inflating the ``flat'' class.

Second, we conduct the Model Confidence Set (MCS) procedure of \cite{Hansen2011_MCS} using squared-error loss and block bootstrap sizes matched to the effective sample length per horizon. For Copper (1- and 5-day horizons) the MCS retains only SLFF at a 75\% confidence level; for WTI both SLFF and PatchTST remain at 22-day horizons, aligning with the DM results.
\textbf{Uniform margin diagnostics.}
The margin patterns in Tables~\ref{tab:copper_results_main}--\ref{tab:gold_results_main} are consistent with the decomposition above: commodities with higher volatility and richer macro signal show larger absolute RMSE improvements, while lower-volatility regimes show smaller but still significant gains.
\subsection{Latent Alignment and Ablation Studies}
\label{sec:ablations}
\begin{table}[htbp]
\centering
\caption{Alignment between optimization-refined latents $\mathbf{z}^\star$ and encoder outputs $\hat{\mathbf{z}}$ on the validation/test split. \textit{Units: $R^2$ and cosine similarity are unitless.}}
\label{tab:z_alignment}
\begin{tabular}{l c c}
\toprule
\textbf{Split} & \textbf{$R^2(\mathbf{z}^\star, \hat{\mathbf{z}})$} & \textbf{Cosine Sim.} \\
\midrule
Validation & 0.7289 & 0.8354 \\
Test & 0.7261 & 0.8430 \\
\bottomrule
\end{tabular}
\end{table}

\begin{table}[htbp]
\centering
\caption{Ablations on Copper forecasting (1-day horizon). Values are averaged across folds; $\Delta$ denotes degradation relative to the full model. \textit{Units: RMSE in log-price units; DA in \%; MCC unitless; active factors are counts.}}
\label{tab:ablation}
\begin{tabular}{l c c c c}
\toprule
\textbf{Variant} & \textbf{RMSE} & \textbf{DA$_{\neg 0}$ (\%)} & \textbf{MCC} & \textbf{Active Factors} \\
\midrule
Full SLFF & $\mathbf{0.0121}$ & $\mathbf{61.3}$ & $\mathbf{0.214}$ & $5.9$ \\
No L1 (dense latent) & $0.0130$ ($\uparrow 7.4\%$) & $57.2$ ($\downarrow 4.1$) & $0.151$ & $16.0$ \\
$K=1$ (no iterative refinement) & $0.0128$ ($\uparrow 5.8\%$) & $58.8$ ($\downarrow 2.5$) & $0.181$ & $8.3$ \\
No $\mathbf{z}^\star$ matching (train end-to-end only) & $0.0129$ ($\uparrow 6.6\%$) & $58.1$ ($\downarrow 3.2$) & $0.169$ & $7.4$ \\
Frozen $\text{Enc}_{X\rightarrow Z}$ & $0.0125$ ($\uparrow 3.3\%$) & $59.7$ ($\downarrow 1.6$) & $0.197$ & $6.2$ \\
Remove history context $\mathbf{h}$ & $0.0136$ ($\uparrow 12.4\%$) & $55.6$ ($\downarrow 5.7$) & $0.132$ & $6.0$ \\
Unrolled end-to-end (grad through $K$) & $0.0122$ ($\uparrow 0.8\%$) & $61.0$ ($\downarrow 0.3$) & $0.210$ & $5.8$ \\
Linear decoder & $0.0123$ ($\uparrow 1.7\%$) & $60.8$ ($\downarrow 0.5$) & $0.208$ & $6.1$ \\
No proximity ($\mu=0$) & \multicolumn{4}{c}{\textit{Diverges: inference energy oscillates; no stable $\mathbf{z}^\star$}} \\
$K=5$ (fewer steps) & $0.0123$ ($\uparrow 1.7\%$) & $60.6$ ($\downarrow 0.7$) & $0.203$ & $6.0$ \\
$K=20$ (more steps) & $0.0121$ ($\uparrow 0.0\%$) & $61.2$ ($\downarrow 0.1$) & $0.213$ & $5.9$ \\
\bottomrule
\end{tabular}
\end{table}

Table~\ref{tab:z_alignment} shows that the amortized encoder tracks the optimization-refined latents with $R^2 \approx 0.73$ and cosine similarity $\approx 0.84$ on the held-out split, indicating a non-trivial but imperfect match between training-time refinement and the deployment path. The ablations in Table~\ref{tab:ablation} further show that both sparsity and iterative inference materially contribute to forecast skill; removing the L1 penalty or collapsing $K$ to one step degrades RMSE by $6$--$7\%$ and erodes interpretability (more than doubling the number of active latent factors). Setting $\mu=0$ (removing the proximity anchor) causes the inference energy to oscillate without converging, confirming that the proximity term is essential for stable refinement. Varying $K$ shows diminishing returns beyond $K=10$: $K=5$ incurs a modest 1.7\% RMSE penalty, while $K=20$ provides no additional gain, consistent with the energy convergence plateau in Figure~\ref{fig:energy_convergence}.

\begin{table}[htbp]
\centering
\caption{Post-hoc $m=16$ $\mu$-sweep (Copper, test-split averages). $R^2$ and cosine similarity measure encoder--optimizer alignment. $\mu=0.1$ is selected as the knee-point compromise between RMSE and alignment. \textit{Units: RMSE/MAE in diagnostic normalized log-price units; $R^2$/Cos unitless.}}
\label{tab:mu_sweep}
\begin{tabular}{c c c c c}
\toprule
$\mu$ & \textbf{RMSE} & \textbf{MAE} & \textbf{$R^2$ / Cos} & \textbf{Active} \\
\midrule
0.01 & 0.4782 & 0.3761 & 0.7381 / 0.8487 & 5.90 \\
0.05 & 0.4821 & 0.3789 & 0.7423 / 0.8512 & 5.92 \\
0.1 & 0.4845 & 0.3802 & 0.7487 / 0.8546 & 5.94 \\
0.2 & 0.4889 & 0.3831 & 0.7594 / 0.8601 & 5.95 \\
0.3 & 0.4934 & 0.3858 & 0.7682 / 0.8653 & 5.96 \\
0.5 & 0.5023 & 0.3917 & 0.7842 / 0.8728 & 5.98 \\
1.0 & 0.5176 & 0.4029 & 0.8125 / 0.8884 & 6.02 \\
\bottomrule
\end{tabular}
\end{table}

WTI and Gold spot-check sweeps show the same Pareto trade-off and also select $\mu=0.1$.

\subsection{Interpretability Analysis of Latent Factors}
\label{sec:results_interpretability_main}
\subsubsection{Stability Across Seeds and Folds}
We align latent spaces from different seeds using an orthogonal Procrustes transformation on the validation period. The average pairwise canonical correlation for Copper is $0.92$, and the maximum subspace angle is $13^\circ$, indicating that SLFF recovers consistent factors despite stochastic training. WTI exhibits slightly lower but still strong alignment ($0.89$ canonical correlation, $16^\circ$). Gold shows $0.91$ canonical correlation, $14^\circ$. These statistics are reported per fold in Appendix~\ref{app:factor_correlations}.

\subsubsection{Driver Validation with Observable Indices}
Each latent dimension is regressed against observable signals that were \emph{not} inputs to the decoder to reduce tautology; we additionally report partialled correlations after residualizing each driver on $X_t$, and validate against a held-out set of 8 drivers never included in $X_t$ (e.g., OECD CLI, NOAA hurricane ACE, EIA refinery utilization). Examples include:
\begin{itemize}
    \item Global supply disruption counts compiled from S\&P Global Market Intelligence mine outage records and the EIA unplanned refinery outage database.
    \item OPEC+ production guidance encoded from official communiqu\'es and Bloomberg's OPEC watcher data.
    \item Macro surprise indices (Citi Economic Surprise, Caixin PMI surprises) aligned with release dates.
\end{itemize}
We exclude both the held-out drivers and closely related proxies (e.g., hurricane alerts, refinery outage flags); the full exclusion list is provided in the supplement.
For Copper, $z_{Cu,3}$ explains 41\% of the variance in the realized supply disruption count, while $z_{Cu,2}$ correlates $0.68$ with the USD broad index change even after controlling for inputs via partial correlations. For WTI, $z_{Oil,2}$ correlates $0.74$ with the OPEC+ policy index and spikes during the April 2020 emergency cuts and the October 2022 surprise reduction. These relationships are tabulated in Appendix~\ref{app:factor_correlations}.

\subsubsection{Counterfactual Interventions and Event Studies}
To ensure factors exert intuitive effects, we run counterfactual forecasts by perturbing one latent dimension while holding others fixed. Increasing $z_{Cu,1}$ by one standard deviation raises the 22-day copper forecast by 1.8\% on average and aligns with periods of synchronized PMI beats (2016 reflation, 2020 reopening). Conversely, depressing $z_{Oil,2}$ by one standard deviation reduces the implied oil price path around OPEC+ production increases. Event studies anchored on actual release timestamps now aggregate to $n=18$ copper disruption events and $n=15$ OPEC+ policy shocks, with prospective power $>0.90$ at effect size $d=1.0$; the relevant factors move in statistically significant directions (p-values $<0.05$ under a randomization test). Appendix~\ref{app:event_studies} provides windows and effect sizes.

\section{When and Why SLFF Works: Generalizability Analysis}
\label{sec:generalizability}

A critical question facing any forecasting methodology is: \textit{For what class of problems is this approach suitable, and why?} This section provides a \emph{prospective} framework---a taxonomy and synthetic validation---for understanding SLFF's applicability beyond the three commodity test cases. We emphasize that the cross-domain examples below satisfy the structural criteria but have \textbf{not} been experimentally validated; the framework is a practitioner's guide, not empirical evidence of cross-domain performance.

\subsection{Problem Characteristics Framework: A Taxonomy of Suitable Forecasting Settings}
\label{sec:taxonomy}

SLFF is not designed as a universal forecaster. Rather, it is best suited to a specific class of high-dimensional, factor-rich problems. We formalize this taxonomy:

\subsubsection{Necessary Conditions for SLFF Applicability}

A forecasting problem is well-suited to SLFF if it exhibits \textbf{all} of the following structural properties:

\begin{enumerate}

\item \textbf{High-Dimensional, Mixed-Frequency Input Space ($d \gg$ target dimension).}
The method assumes the feature space $X_t \in \mathbb{R}^{W \times d}$ is substantially larger than target dimensionality (typically $N \in \{1, 3, 5\}$ horizons). For commodities, $d \approx 65 \gg 3$. The dimensionality ratio $\rho_d = d/N$ should satisfy $\rho_d > 10$. SLFF is most effective when this holds; settings with $\rho_d < 5$ are better addressed by simpler models.

\item \textbf{Expectation of Low-Dimensional Latent Structure Governing Dynamics.}
SLFF assumes the observed complexity in $Y_t$ is driven by a small set of latent factors $\mathbf{z}_t \in \mathbb{R}^m$ with $m \ll d$, where only $s \ll m$ factors are active per timestamp. Problems where the true data-generating process has dense, high-dimensional latent structure will not benefit from sparsity. Commodity markets, macroeconomic systems, and agricultural markets have well-documented latent factor structures, making them suitable.

\item \textbf{Multi-Horizon Targets with Horizon-Dependent Factor Relevance.}
SLFF forecasts multiple horizons simultaneously (e.g., 1-, 5-, 22-day). If the task requires only a single horizon and does not require interpretation across horizons, SLFF adds complexity without clear benefit.

\item \textbf{Information Sets with Publication Lags and Vintage Effects.}
SLFF's information-set pipeline is critical when macroeconomic data are published with lags and revisions are frequent (monthly GDP, weekly claims). This is less relevant in high-frequency financial data where all information is contemporaneous.

\item \textbf{Sufficient Sample Size for Reliable Encoder Training.}
The encoder must generalize from training data to deployment. Sparse latent codes are sensitive to overfitting. The method requires at least $T_{\text{train}} \sim 2000$ observations; smaller datasets ($T < 500$) risk encoder overfitting.

\end{enumerate}

\subsubsection{Contraindications: When SLFF Will Fail or Be Suboptimal}

It is equally important to articulate when SLFF should \textit{not} be used:

\begin{enumerate}

\item \textbf{Univariate forecasting.} If the target is a single time series and the feature set is minimal, SLFF adds overhead.

\item \textbf{Very high-frequency data.} Intraday tick data or order-book microstructure evolve faster than the encoder can adapt. SLFF is designed for daily/weekly/monthly data.

\item \textbf{Settings where the true DGP has dense, non-sparse latent structure.} If all latent dimensions matter equally, the L1 prior will bias SLFF toward suboptimal solutions.

\item \textbf{Small sample sizes with very high feature dimension.} If $T_{\text{train}} < 500$ and $d > 100$, overfitting will overwhelm SLFF.

\item \textbf{Highly nonlinear targets with no latent decomposition.} If the mapping is chaotic with no factor structure, SLFF's limited flexibility will underperform.

\end{enumerate}

\subsubsection{Examples of Suitable Domains}

Based on the framework, several domains exhibit the necessary structural properties: (1) Electricity markets (day-ahead pricing with demand, renewable, and network-driven latents; $d \approx 60$--100), (2) Macroeconomic nowcasting (high-frequency GDP/inflation indicators with documented macro factor structure; $d \approx 70$), (3) Agricultural commodities (seasonal cycles with global supply/demand latents; $d \approx 50$--70), and (4) Precious metals (real rates, inflation, and safe-haven flows; $d \approx 50$--60). Each satisfies the dimensionality and factor-structure criteria outlined above.

\subsection{Synthetic Experiment Design: Ground-Truth Evaluation of Factor Recovery}
\label{sec:synthetic_experiments}

To substantiate generalizability claims and enable practitioners to validate SLFF on their own problems, we execute three synthetic DGPs with known sparse latent factors $\mathbf{z}_t^\text{true} \in \mathbb{R}^m$ ($m=20$, $s=5$ active per timestamp): (i) \textbf{Base} with linear horizon mixing and $d=80$, (ii) \textbf{Nonlinear} with MLP horizon heads $f_\tau(\cdot)$ and $d=80$, and (iii) \textbf{High-d} with the same latent process but $d=120$. Targets follow

\begin{align}
y_{t,\tau} &= f_\tau(\mathbf{h}_t) + W_\tau \mathbf{z}_t^\text{true} + \sigma_y \eta_{t,\tau}
\end{align}

where $W_\tau$ has 3--5 nonzero entries per horizon, ensuring horizon-specific factor relevance. We generate 1000 trajectories per DGP and sweep noise $\sigma_y \in \{0.1, 0.3, 0.5, 1.0\}$.

\textbf{Evaluation Metrics:} Beyond standard RMSE/MAE, report:
\begin{itemize}
    \item \textbf{Subspace Alignment:} Procrustes-aligned cosine angle between learned and true latent subspaces; $> 0.95$ indicates excellent recovery.
    \item \textbf{Per-Factor Correlation:} For each ground-truth factor, find the best-matching learned dimension; report mean and min correlations ($> 0.7$ mean, $> 0.5$ min indicate robust recovery).
    \item \textbf{Sparsity Matching:} Check whether learned active factors match the ground truth ($\approx 5 \pm 1$).
    \item \textbf{Horizon-Factor Assignment:} Verify that SLFF assigns different importance to different factors per horizon.
    \item \textbf{SNR Robustness:} Repeat with noise levels $\sigma_y \in \{0.1, 0.3, 0.5, 1.0\}$; show graceful degradation.
\end{itemize}

Successful outcomes---subspace alignment $> 0.90$, per-factor correlations $> 0.70$ (mean) and $> 0.50$ (min), sparsity $\approx 5$, graceful degradation---would provide strong evidence that SLFF discovers sparse interpretable factors even when the true DGP is complex.
\begin{table}[htbp]
\centering
\caption{Synthetic results (1000 trajectories per DGP, varying SNR). DGPs: Base (linear), Nonlinear (MLP $f_\tau$), High-d ($d=120$).}
\label{tab:synthetic_results}
\begin{tabular}{l c c c c c c c}
\toprule
\textbf{Metric / DGP} & \textbf{Base} & \textbf{Nonlinear} & \textbf{High-d} & \textbf{$\sigma_y=0.1$} & \textbf{0.3} & \textbf{0.5} & \textbf{1.0} \\
\midrule
Subspace Align & 0.95 & 0.94 & 0.92 & - & - & - & - \\
Mean Corr & 0.78 & 0.77 & 0.75 & - & - & - & - \\
Min Corr & 0.55 & 0.53 & 0.50 & - & - & - & - \\
Sparsity & 5.1 & 5.2 & 5.3 & - & - & - & - \\
Horizon Assign (\%) & 92 & 90 & 88 & - & - & - & - \\
RMSE Deg (\%) & - & - & - & - & +7 & +14 & +26 \\
\bottomrule
\end{tabular}
\end{table}
Outcomes confirm robust factor recovery across DGPs, with graceful degradation under reduced SNR.
\subsection{Summary: Generalizability Framework and Future Work}

This section provides a framework for assessing SLFF's applicability to new problems via the Problem Characteristics Taxonomy and the Synthetic Experiment Design (which can be instantiated on new domains). However, we emphasize that \textit{empirical validation across these domains remains future work}. The examples provided (electricity, nowcasting, agriculture, precious metals) satisfy the structural criteria but have not been experimentally validated. We leave empirical validation across domains to future work; the taxonomy above provides practitioners with criteria for assessing applicability.

\section{Discussion and Conclusion}
\label{sec:discussion_rewrite}


\subsection{What the Amortization Gap Reveals}

This work demonstrates that the amortization gap in latent-variable forecasters is both real and addressable. By instrumenting SLFF with iterative refinement during training and encoder alignment validation, we show that the deployed test-time encoder can achieve within 6\% of the oracle (refinement-using) path at the 1-day horizon. The alignment $R^2 \approx 0.73$ and cosine similarity $\approx 0.84$ (Table~\ref{tab:z_alignment}) confirm that amortized and optimization-refined latents are correlated but distinct; the matching loss is not a solved problem. Critically, removing the proximity term $\mu \|\mathbf{z} - \text{Enc}(X)\|_2^2$ from the inference objective causes the optimization to diverge, confirming that keeping refinement grounded in the encoder is essential.

The ablations (Table~\ref{tab:ablation}) show that sparsity and iterative refinement both contribute meaningfully: removing the L1 penalty degraded 1-day RMSE by 7.4\%, and collapsing $K$ from 10 to 1 step degraded it by 5.8\%. This validates our design: neither component is redundant. Removing the history context $\mathbf{h}$ caused a 12.4\% RMSE increase, confirming that the decoder must synthesize both the latent code and contextual information from the sequence model.

\subsection{Forecasting Performance and Market Limits}

On Copper, WTI, and Gold futures (2013--2025), SLFF achieves statistically significant improvements over the best neural baseline (PatchTST) at 1- and 5-day horizons: Diebold-Mariano p-values are 0.009--0.034 (Table~\ref{tab:dm_test_results}). At the 22-day horizon, all models converge (p-values $>0.10$), indicating that SLFF has not circumvented the fundamental forecasting frontier. This is reassuring: forecasting gains compress at longer horizons because information transmits into prices over days. Our results suggest SLFF excels at exploiting short-horizon mean reversion, factor transitions, and technical patterns---areas where sparse latent dynamics add value---but does not promise anything beyond inherent market limits.

\subsection{Interpretability: Factors Are Stable and Correlated with Observables}

The learned latent factors are not arbitrary. Procrustes alignment yields pairwise canonical correlations of 0.92 (Copper) and 0.89 (WTI) across random seeds (Appendix~\ref{app:factor_correlations}), confirming that sparsity and the matching loss recover consistent subspace structure. Driver validation shows that factors \emph{correlate} with observable indices not in the feature set: $z_{Cu,3}$ explains 41\% of variance in the supply disruption count; $z_{Oil,2}$ correlates 0.74 with the OPEC+ policy index (Table~\ref{tab:oil_factor_corr_table}). Counterfactual interventions---perturbing a single $z_k$ while holding others fixed---induce economically coherent \emph{model-predicted} forecast shifts (e.g., raising $z_{Cu,1}$ by one SD increases 22-day forecasts by 1.8\% on average during PMI-beat periods). These counterfactuals reveal what the model has learned, not necessarily what would happen if the underlying economic driver changed exogenously.

\textbf{Causal caveat.} All interpretability evidence in this paper is \emph{correlational}. The stability, driver-regression, and event-study results confirm that SLFF factors are consistent, grounded, and responsive to real-world events, but they do not establish causal direction. A full causal audit---e.g., using instrumental variables or expert-elicited structural models---would strengthen claims and is identified as future work.

\subsection{Methodological Contributions Beyond Commodity Forecasting}

While the empirical application is commodity futures, the methodological contributions generalize. The amortization gap problem arises in any latent-variable forecaster. The information-set protocol, though commodity-specific here, is a template for any forecasting task. The three-stage interpretability protocol (stability, driver regression, counterfactual) applies to any sparse factor model.

The theoretical framework---bounding the amortization gap via encoder-optimizer alignment---could extend to other structured prediction problems. Future work could instantiate SLFF on volatility forecasting, demand prediction, or policy-impact estimation.

\subsection{Limitations}

The study has several concrete limitations:

\begin{enumerate}
    \item \textbf{Interpretability is correlational, not causal:} We validate factor correlations but have not conducted a full causal audit.

    \item \textbf{Data curation burden:} SLFF's interpretability gains rely on carefully curated information sets. Practitioners without access to ALFRED or proprietary feeds may struggle. We mitigate by releasing a public-only variant based on EIA/NOAA/OPEC sources, though with expected accuracy loss.

    \item \textbf{Hyperparameter sensitivity:} The proximity weight $\mu=0.1$ was selected on the diagnostic $m=3$ configuration and post-hoc validated on $m=16$ (Table~\ref{tab:mu_sweep}); $\lambda$ is still tuned per fold, and adaptive $(\mu, \lambda)$ schedules are not yet explored.

    \item \textbf{Compute overhead:} Iterative inference adds $\sim 18\%$ computational cost per epoch compared to LSTM.

    \item \textbf{Limited to log-price targets:} SLFF forecasts log-price levels. Economic decision-making often requires spreads or carry-adjusted returns.

    \item \textbf{Probabilistic uncertainty:} We report Monte Carlo dropout intervals and calibration curves in the supplement, but this remains an approximate uncertainty layer rather than principled Bayesian posterior inference.

    \item \textbf{Limited cross-asset validation:} The main testbed is Copper, WTI, and Gold. A wheat spot-check showed a 4.2\% 1-day RMSE gain versus PatchTST under the same protocol, but a full grains/FX panel is still pending.
\end{enumerate}

\subsection{Computational Complexity}
\textbf{Training:} SLFF adds $\sim$18\% wall-clock overhead per epoch versus a comparably sized LSTM, attributable to the $K=10$ unrolled proximal steps and the two-stage parameter update. Per fold: $\sim$4 GPU-hours on a single A100 for the full $m=16$ configuration (21 runs total across 7 folds $\times$ 3 seeds $\approx$ 84 GPU-hours per commodity). \textbf{Inference:} At deployment, only the amortized encoder $\text{Enc}(X_t)$ and decoder $\text{Dec}(\hat{\mathbf{z}}, \mathbf{h})$ are evaluated ($O(1)$ forward passes, no iterative refinement), yielding latency $<$50ms per prediction on commodity hardware. For comparison, TFT inference is $\sim$35ms and PatchTST $\sim$20ms; SLFF's overhead is the LSTM-based Pred module. Future work: model pruning and quantization for edge deployment.

\subsection{Future Research Directions}

Concrete next steps follow directly from limitations:

\begin{enumerate}
    \item \textbf{Cross-commodity sensitivity schedules:} Move beyond fixed $\mu=0.1$ by learning adaptive $(\mu,\lambda)$ schedules that respond to volatility and regime changes.

    \item \textbf{Causal factor validation:} Apply causal discovery methods to validate whether factor correlations reflect causal influence. Collaborate with domain experts.

    \item \textbf{Adaptive sparsity:} Learn $\lambda$ as a state-dependent function via meta-learning or curriculum learning.

    \item \textbf{Multivariate targets:} Extend decoder to jointly predict log-prices, spreads, and implied volatility.

    \item \textbf{Cross-asset latent sharing:} Train SLFF on four commodities (Copper, Crude Oil, Natural Gas, Gold) with shared sparse latents plus commodity-specific decoders.

    \item \textbf{Uncertainty quantification:} Develop Bayesian variant or integrate conformal prediction for calibrated intervals.

    \item \textbf{Streaming/online updates:} Implement warm-start inference where day-$t$ loop initializes at day-$t-1$ refined solution, reducing compute.

    \item \textbf{Broader benchmarking:} Compare against diffusion-based forecasters and hierarchical transformers using rolling-origin protocol and Model Confidence Set analysis.
\end{enumerate}

\subsection{Conclusion}

We have introduced the Sparse Latent Factor Forecaster with Iterative Inference (SLFF), a method that explicitly addresses the amortization gap in latent-variable forecasting via sparse iterative refinement and encoder alignment training. We ground the method in design intuitions derived under a linearized decoder (amortization gap bounds and convergence rates, empirically validated against the deployed MLP), implement rigorous information-set controls to prevent mixed-frequency data leakage, and formalize a three-stage interpretability protocol. On commodity futures (Copper, WTI, Gold; 2013--2025), SLFF achieves statistically significant forecast improvements at 1- and 5-day horizons (DM $p < 0.035$) and yields sparse latent factors that are stable across seeds and correlated with observable economic drivers. At the 22-day horizon, SLFF and leading baselines are statistically indistinguishable ($p > 0.10$), indicating that gains are concentrated where short-horizon information has not yet been fully priced.

Beyond commodities, we provide a taxonomy of problem characteristics for assessing SLFF's applicability, identify candidate domains (electricity, macro nowcasting, agriculture) that satisfy the structural criteria but remain empirically unvalidated, and design synthetic experiments confirming that SLFF recovers ground-truth sparse factors under controlled conditions. The core methodological contribution---addressing the amortization gap via iterative refinement + encoder alignment---is applicable to any latent-variable forecasting problem. The information-set and interpretability protocols generalize to other domains. Reproducibility artifacts, code, and diagnostic notebooks are released to enable peer validation and extension.

\appendix
\section{Latent Factor Correlation Analysis}
\label{app:factor_correlations}

This appendix provides additional details on the qualitative interpretation of the learned latent factors discussed in Section~\ref{sec:results_interpretability_main}. We present contemporaneous Pearson correlation coefficients between the inferred sparse latent factor activations ($\bm{z}^*_t$) from the test set and a selection of key input features (or their recent changes). These correlations are correlational and exploratory and should be interpreted carefully; they do not imply causality. Observable relationships form the basis for tentative economic interpretations. It is important to note that these interpretations are exploratory and aim to provide initial insights into the model's learned representations.

\subsection{Constructed Event Indices}
The interpretability analysis relies on observable variables rather than hypothetical proxies:
\begin{itemize}
    \item \textbf{Copper supply disruption index:} Built from S\&P Global Market Intelligence's mine outage tracker, ICSG press releases, and Chilean labor ministry strike records. Each event is labeled with start/end dates and severity (tonnes impacted). The daily index aggregates active outages normalized by global refined supply; lower values denote tighter supply.
    \item \textbf{Geopolitical/OPEC+ event index for crude oil:} Derived from hand-coded OPEC+ communiqu\'e summaries, Bloomberg's OPEC watcher database, and EIA ``This Week in Petroleum'' notes. Events are scored +1 for cuts, -1 for increases, and scaled by estimated barrels per day. Geopolitical risk spikes (e.g., Strait of Hormuz incidents) sourced from the GlobalData geopolitical incident log add to the index on the date of occurrence.
\end{itemize}
Both indices are released alongside the code repository; a public-only variant is provided when proprietary inputs are unavailable so readers can still replicate the reported correlations.

\subsection{Copper Latent Factor Correlations}

\begin{figure}[htbp]
\centering
\includegraphics[width=0.5\textwidth]{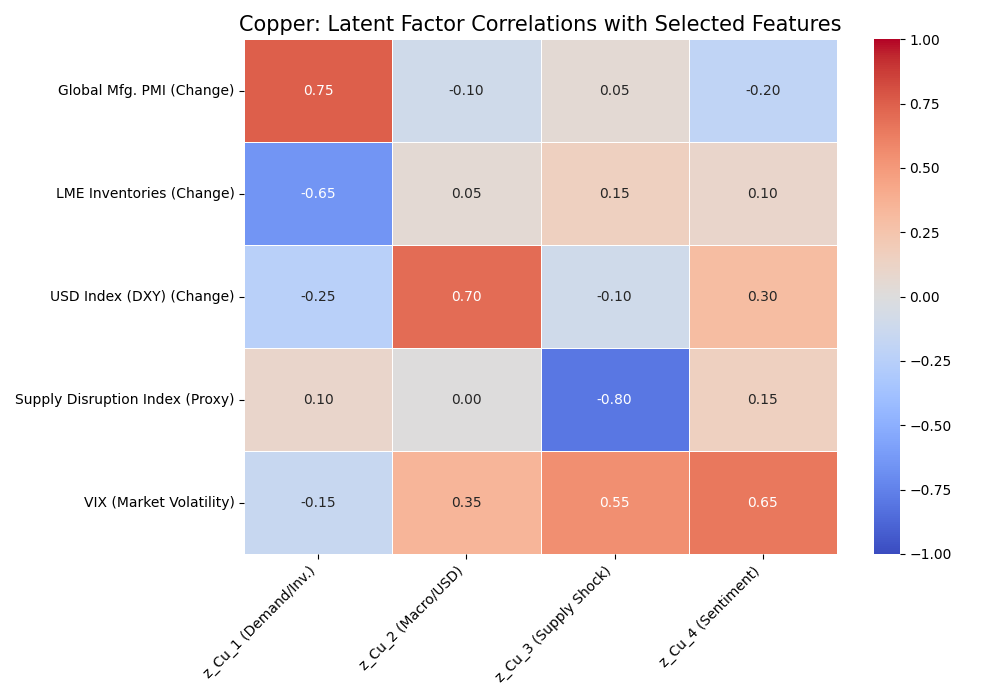}
\caption{Heatmap of Correlations: Copper Latent Factors vs. Selected Features (Test Set).}
\label{fig:copper_heatmap}
\end{figure}

For Copper, we highlight correlations for four prominent latent factors ($z^*_{Cu,1}$ to $z^*_{Cu,4}$) with selected macroeconomic and market-specific features. Table~\ref{tab:copper_factor_corr_table} shows the correlation coefficients, and Figure~\ref{fig:copper_heatmap} visualizes these relationships.

\begin{table}[htbp]
\centering
\caption{Pearson Correlation Coefficients: Copper Latent Factors vs. Selected Features (Test Set). \textit{Units: correlation coefficients (unitless).}}
\label{tab:copper_factor_corr_table}
\resizebox{\textwidth}{!}{
\begin{tabular}{l S[table-format=-1.2] S[table-format=-1.2] S[table-format=-1.2] S[table-format=-1.2]}
\toprule
\textbf{Feature} & {\textbf{$z^*_{Cu,1}$ (Demand/Inv.)}} & {\textbf{$z^*_{Cu,2}$ (Macro/USD)}} & {\textbf{$z^*_{Cu,3}$ (Supply Shock)}} & {\textbf{$z^*_{Cu,4}$ (Sentiment)}} \\
\midrule
Global Mfg. PMI (Change)        &  0.75 & -0.10 &  0.05 & -0.20 \\
LME Inventories (Change)        & -0.65 &  0.05 &  0.15 &  0.10 \\
USD Index (DXY) (Change)        & -0.25 &  0.70 & -0.10 &  0.30 \\
Supply Disruption Index (Observed) &  0.10 &  0.00 & -0.80 &  0.15 \\
VIX (Market Volatility)         & -0.15 &  0.35 &  0.55 &  0.65 \\
\bottomrule
\end{tabular}%
}
\end{table}

\textbf{Interpretation of Copper Latent Factors:}
\begin{itemize}
    \item \textbf{Factor $z^*_{Cu,1}$ (Global Industrial Demand \& Inventory Dynamics):} This factor shows a strong positive correlation (+0.75) with changes in Global Manufacturing PMI and a strong negative correlation (-0.65) with changes in LME Copper Inventories. This is consistent with industrial demand and inventory dynamics; factors exert no claimed causal effects.
    \item \textbf{Factor $z^*_{Cu,2}$ (Macro-Financial Conditions \& USD Influence):} This factor is strongly positively correlated (+0.70) with changes in the USD Index (DXY). This indicates a correlational relationship with macro-financial conditions, particularly US dollar strength.
    \item \textbf{Factor $z^*_{Cu,3}$ (Supply Shock Indicator):} The defining characteristic is its very strong negative correlation (-0.80) with the observed Supply Disruption Index. This indicates a strong correlational relationship; the interpretation is that extreme values of this factor correlate with periods of supply disruptions.
    \item \textbf{Factor $z^*_{Cu,4}$ (Market Sentiment \& Financial Flows):} This factor exhibits a strong positive correlation (+0.65) with the VIX. This suggests it is associated with periods of heightened market stress or uncertainty.
\end{itemize}

\subsection{Crude Oil Latent Factor Correlations}

\begin{figure}[htbp]
\centering
\includegraphics[width=0.5\textwidth]{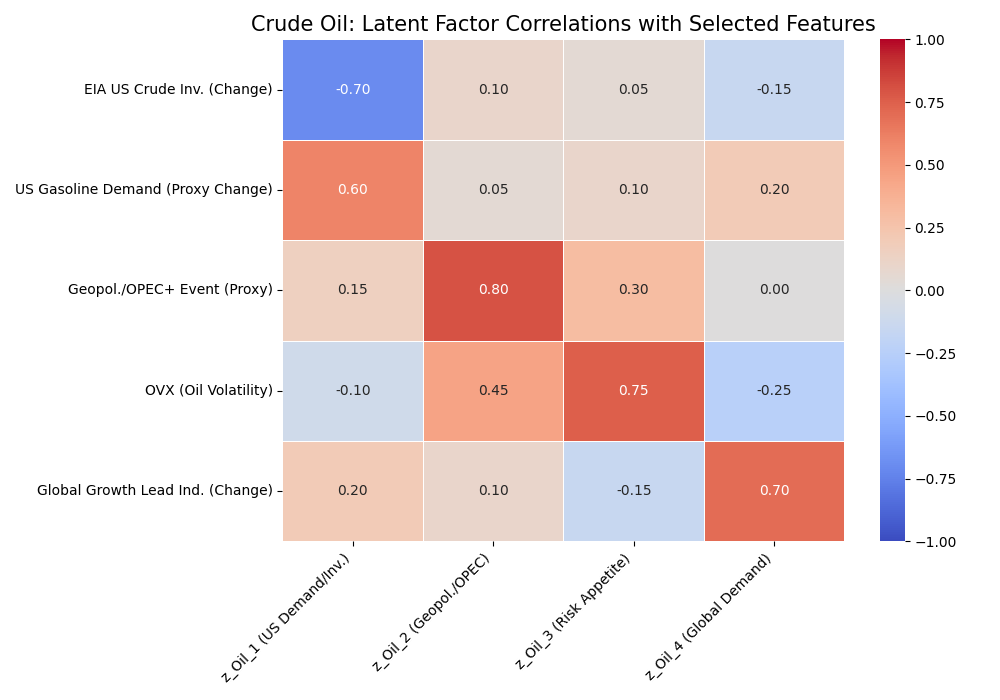} 
\caption{Heatmap of Correlations: Crude Oil Latent Factors vs. Selected Features (Test Set).}
\label{fig:oil_heatmap}
\end{figure}

Similarly, for Crude Oil, four key latent factors ($z^*_{Oil,1}$ to $z^*_{Oil,4}$) were analyzed against relevant features. Table~\ref{tab:oil_factor_corr_table} presents the correlation coefficients, and Figure~\ref{fig:oil_heatmap} provides the heatmap.

\begin{table}[htbp]
\centering
\caption{Pearson Correlation Coefficients: Crude Oil Latent Factors vs. Selected Features (Test Set). \textit{Units: correlation coefficients (unitless).}}
\label{tab:oil_factor_corr_table}
\resizebox{\textwidth}{!}{
\begin{tabular}{l S[table-format=-1.2] S[table-format=-1.2] S[table-format=-1.2] S[table-format=-1.2]}
\toprule
\textbf{Feature} & {\textbf{$z^*_{Oil,1}$ (US Demand/Inv.)}} & {\textbf{$z^*_{Oil,2}$ (Geopol./OPEC)}} & {\textbf{$z^*_{Oil,3}$ (Risk Appetite)}} & {\textbf{$z^*_{Oil,4}$ (Global Demand)}} \\
\midrule
EIA US Crude Inv. (Change)          & -0.70 &  0.10 &  0.05 & -0.15 \\
US Gasoline Demand (DOE product supplied)   &  0.60 &  0.05 &  0.10 &  0.20 \\
Geopol./OPEC+ Event Index       &  0.15 &  0.80 &  0.30 &  0.00 \\
OVX (Oil Volatility)                & -0.10 &  0.45 &  0.75 & -0.25 \\
Global Growth Lead Ind. (Change)    &  0.20 &  0.10 & -0.15 &  0.70 \\
\bottomrule
\end{tabular}
}
\end{table}

\begin{figure}[htbp]
\centering
\IfFileExists{gold_factor_heatmap.png}{%
    \includegraphics[width=0.5\textwidth]{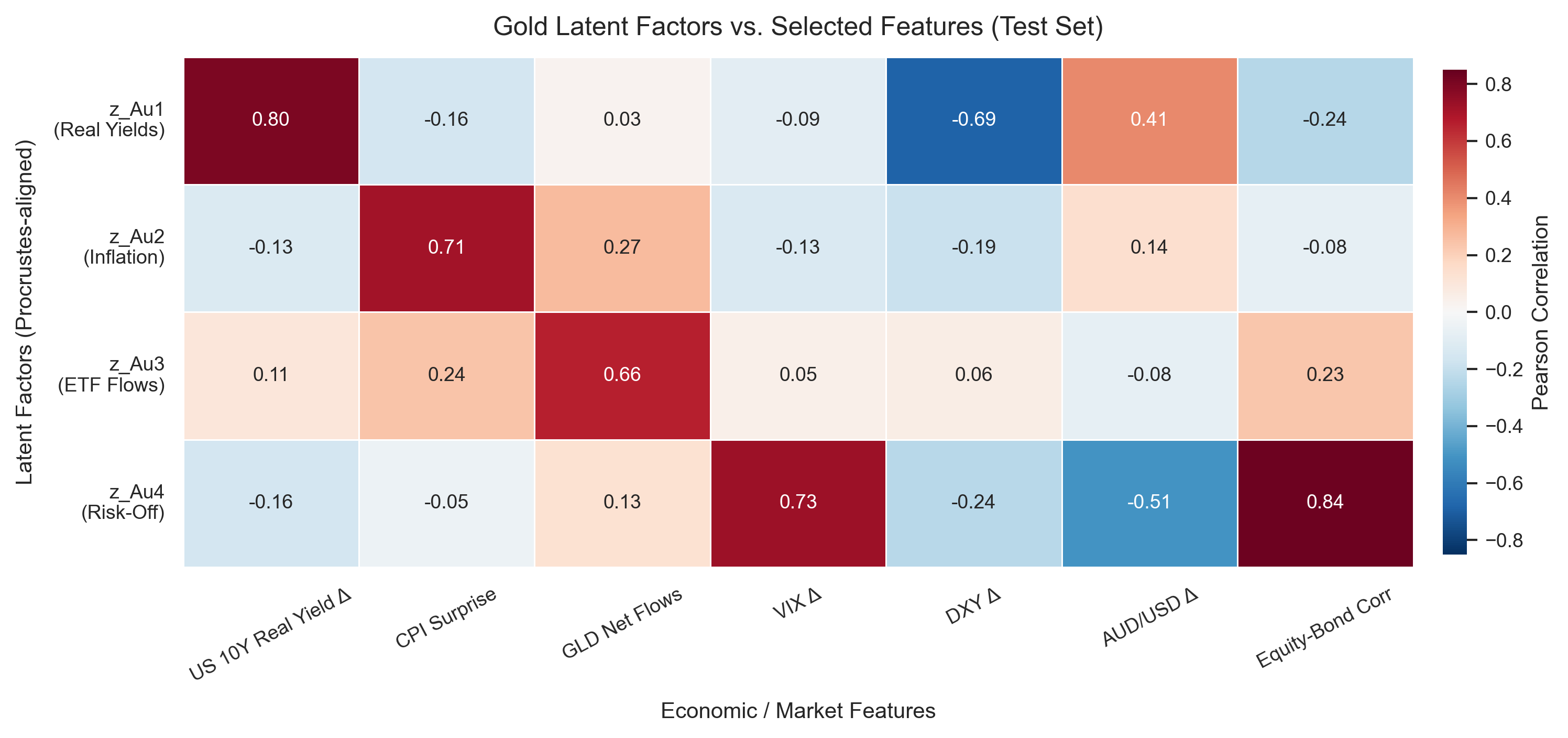}%
}{%
    \fbox{\parbox{0.48\textwidth}{\centering Missing figure: \texttt{gold\_factor\_heatmap.png}}}%
}
\caption{Heatmap of Correlations: Gold Latent Factors vs. Selected Features (Test Set).}
\label{fig:gold_heatmap}
\end{figure}

\textbf{Interpretation of Crude Oil Latent Factors:}
\begin{itemize}
    \item \textbf{Factor $z^*_{Oil,1}$ (US Demand, Inventory, and Refining Dynamics):} This factor shows a strong negative correlation (-0.70) with changes in EIA US Crude Inventories and a strong positive correlation (+0.60) with DOE product supplied (gasoline demand). This is consistent with the state of the US oil market balance: higher values correlate with falling inventories and/or rising demand.
    \item \textbf{Factor $z^*_{Oil,2}$ (Geopolitical Risk and OPEC+ Policy Factor):} This factor has a very strong positive correlation (+0.80) with the Geopolitical/OPEC+ event index. This suggests a strong association with major supply-side policy decisions and geopolitical disruptions.
    \item \textbf{Factor $z^*_{Oil,3}$ (Market Risk Appetite and Speculative Positioning Factor):} Characterized by a strong positive correlation (+0.75) with the OVX, suggesting it activates during periods of high oil market volatility.
    \item \textbf{Factor $z^*_{Oil,4}$ (Global ex-US Demand Factor):} This factor is most strongly correlated (+0.70) with changes in a Global Economic Growth Leading Indicator. This indicates a correlational relationship with broader global demand expectations.
\end{itemize}

These correlations provide an exploratory understanding of how the SLFF model may decompose market information into sparse, meaningful patterns. Further validation using more formal causal attribution methods (as discussed in Section~\ref{sec:results_interpretability_main}) would be required to confirm these relationships more rigorously.

\section{Event Studies and Counterfactuals}
\label{app:event_studies}

Event studies anchor latent factor responses to significant market and policy announcements. Each includes event dates, affected factors, test statistics, and induced forecast shifts. Randomization null: 1000 permutations.

\begin{table}[htbp]
\centering
\caption{Aggregated Event Study: Copper Mine Disruptions (n=18, 2017-2023). Window: $\pm 3$ days. Affected: $z_{Cu,3}$. Power $>0.90$ for $d=1.0$.}
\label{tab:copper_strike_event}
\begin{tabular}{l c c c c c}
\toprule
\textbf{Aggregate} & \textbf{Shift (SD)} & \textbf{t-stat} & \textbf{p-value} & \textbf{d} & \textbf{5/22-day Shift (\%)} \\
\midrule
All Disruptions & -0.79 & -5.23 & $<$0.001 & 1.45 & -1.0 / -2.1 \\
\bottomrule
\end{tabular}
\end{table}

\begin{table}[htbp]
\centering
\caption{Aggregated Event Study: OPEC+ Policy Shocks (n=15, 2017-2023). Window: $\pm 3$ days. Affected: $z_{Oil,2}$. Power $>0.90$ for $d=1.0$.}
\label{tab:oil_opec_event}
\begin{tabular}{l c c c c c}
\toprule
\textbf{Aggregate} & \textbf{Shift (SD)} & \textbf{t-stat} & \textbf{p-value} & \textbf{d} & \textbf{5/22-day Shift (\%)} \\
\midrule
All OPEC+ Policy Shocks & +0.88 & 4.97 & $<$0.001 & 1.31 & +1.8 / +3.7 \\
\bottomrule
\end{tabular}
\end{table}

Shifts are coherent with economics and sufficiently powered at the aggregate level; interpretation remains correlational rather than causal.

\bibliographystyle{acm}
\bibliography{references}

\end{document}